\documentclass{article}

\usepackage{arxiv}
 
\usepackage[utf8]{inputenc} 
\usepackage[T1]{fontenc}    
\usepackage[colorlinks=true, urlcolor=blue, linkcolor=black, citecolor=blue, anchorcolor=black, pdfborder={0 0 0}]{hyperref}         
\usepackage{url}            
\usepackage{booktabs}       
\usepackage{amsfonts}       
\usepackage{nicefrac}       
\usepackage{microtype}      
\usepackage{lipsum}
\usepackage{graphicx}
\graphicspath{ {./images/} }
 
\usepackage{wrapfig}
\usepackage{graphicx}
\usepackage{booktabs}
\usepackage{multirow}
\usepackage{amssymb}
\usepackage{pifont}
\usepackage{tikz}
\usepackage{caption}
\usepackage{multicol}
\usepackage{adjustbox}
\usepackage[utf8]{inputenc}
\usepackage[greek,english]{babel}
\usepackage{subcaption}
\usepackage[sort&compress,authoryear]{natbib}
\usepackage{amsmath}
\usepackage[capitalize,nameinlink]{cleveref}
 
\newcommand{\ie}{i.e.\ }
\newcommand{\eg}{e.g.\ }

\usepackage{floatrow}
\usepackage{array}
 
\newcommand{\cmark}{\checkmark}
\newcommand{\xmark}{\ding{55}}
 
\usepackage[normalem]{ulem}
\newcommand\added[1]{#1}        
\newcommand\removed[1]{}        

\begin{document}

\title{Thalia: A Global, Multi-Modal Dataset for Volcanic Activity Monitoring}

\author{
 Nikolas Papadopoulos\hspace{0.5mm}\href{https://orcid.org/0009-0001-5784-682X}{\includegraphics[height=2.0ex]{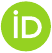}}$^{1,*}$ \\
  \And
 Nikolaos Ioannis Bountos\hspace{0.5mm}\href{https://orcid.org/0000-0003-1615-0196}{\includegraphics[height=2.0ex]{orcid.pdf}}$^{1,2}$ \\
  \And
 Maria Sdraka\hspace{0.5mm}\href{https://orcid.org/0000-0003-2053-1274}{\includegraphics[height=2.0ex]{orcid.pdf}}$^{1,2}$ \\
  \AND
 Andreas Karavias\hspace{0.5mm}\href{https://orcid.org/0000-0002-3284-5134}{\includegraphics[height=2.0ex]{orcid.pdf}}$^{1}$ \\
  \And
 Gustau Camps-Valls\hspace{0.5mm}\href{https://orcid.org/0000-0003-1683-2138}{\includegraphics[height=2.0ex]{orcid.pdf}}$^{3}$ \\
  \And
 Ioannis Papoutsis\hspace{0.5mm}\href{https://orcid.org/0000-0002-2845-9791}{\includegraphics[height=2.0ex]{orcid.pdf}}$^{1,4}$ \\
}

\maketitle

\begin{center}
\small
$^{1}$Remote Sensing Lab, National Technical University of Athens, Greece \\
$^{2}$Department of Informatics and Telematics, Harokopio University of Athens, Greece \\
$^{3}$Image Processing Laboratory (IPL), Universitat de Val\`encia, Spain \\
$^{4}$National Observatory of Athens, Greece \\
$^{*}$Corresponding author: \texttt{npapadopoulos@mail.ntua.gr}
\end{center}

\begin{center}
\end{center}

\begin{abstract}
Monitoring volcanic activity is of paramount importance to safeguarding lives, infrastructure, and ecosystems. However, only a small fraction of known volcanoes are continuously monitored. 
Satellite-based Interferometric Synthetic Aperture Radar (InSAR) enables systematic, global-scale deformation monitoring. However, its complex data challenge traditional remote sensing methods. Deep learning offers a powerful means to automate and enhance InSAR interpretation, advancing volcanology and geohazard assessment. Despite its promise, progress has been limited by the scarcity of well-curated datasets. In this work, we build on the existing \textit{Hephaestus} dataset and introduce \textit{Thalia}, addressing crucial limitations and enriching its scope with higher-resolution, multi-source, and multi-temporal data. \textit{Thalia} is a global collection of 38 spatiotemporal datacubes covering 7 years and integrating InSAR products, topographic data, as well as atmospheric variables, known to introduce signal delays that can mimic ground deformation in InSAR imagery. Each sample includes expert annotations detailing the type, intensity, and extent of deformation, accompanied by descriptive text. To enable fair and consistent evaluation, we provide a comprehensive benchmark using state-of-the-art models for classification and segmentation. This work fosters collaboration between machine learning and Earth science, advancing volcanic monitoring and promoting automated, deep-learning based approaches in geoscience. The code and latest version of the dataset are available through the github repository: \url{https://github.com/Orion-AI-Lab/Thalia}.
\end{abstract}

\begin{figure*}
    \centering
    \includegraphics[width=\linewidth]{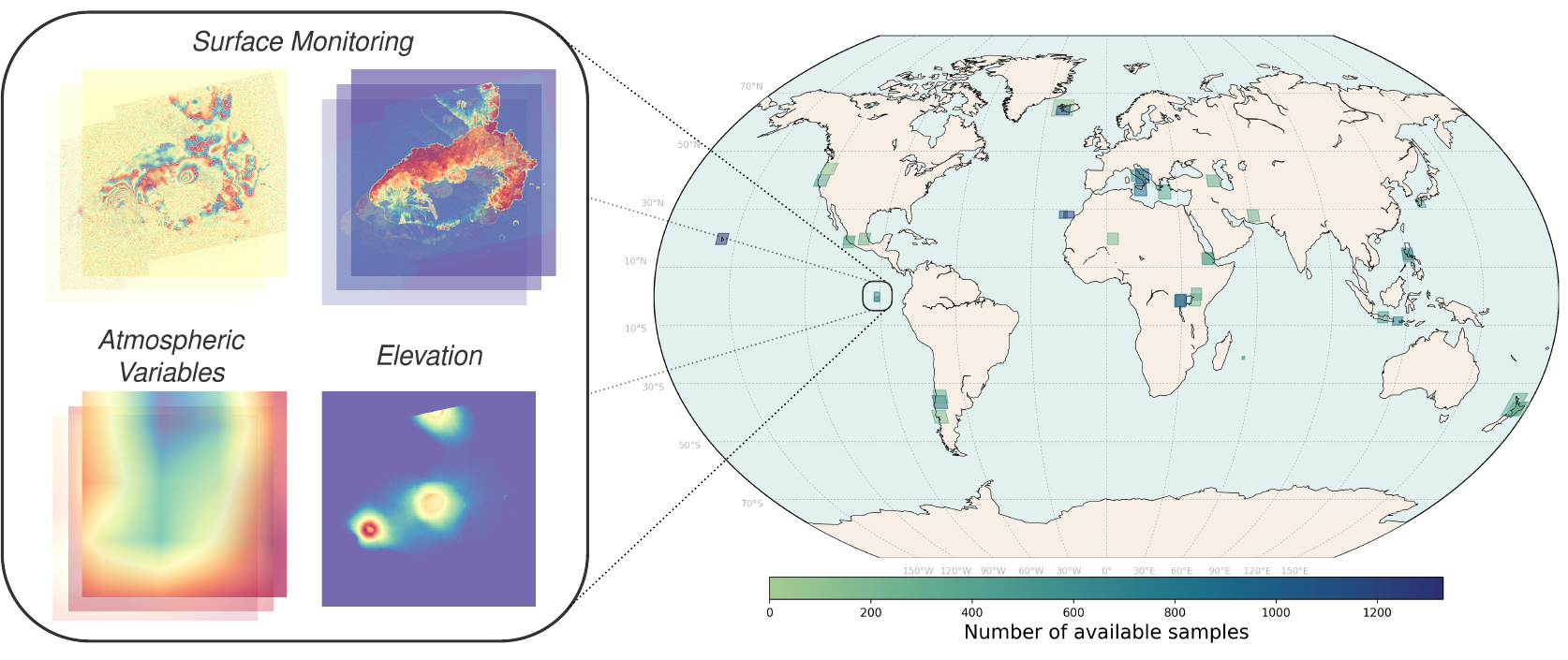}
    \caption{High-level overview of the \textit{Thalia} dataset. \textbf{Left:} The data sources integrated in each spatiotemporal datacube, illustrated in four panels: \textit{Surface Monitoring} (InSAR wrapped phase difference and coherence), \textit{Atmospheric Variables} (total column water vapor, surface pressure, and vertically integrated temperature), and \textit{Elevation} (Digital Elevation Model). \textbf{Right:} Global distribution of the 38 datacubes. Box sizes reflect the spatial extent of each processing frame and color intensity the number of available InSAR products.}
    \label{fig:main}
\end{figure*}

\section{Introduction}
\label{sec:introduction}
Monitoring volcanic activity is essential for protecting lives, infrastructure, and ecosystems \citep{marti2024rest}. Yet, only a small fraction of the world’s volcanoes are continuously observed. 
Ground deformation monitoring plays a vital role in volcanic hazard assessment, providing early insights into subsurface magmatic activity~\citep{https://doi.org/10.1029/2001RG000107}. It constitutes one of the most reliable eruption precursors, with signals that can appear days to even years before an event \citep{biggs2014global}. Timely detection of such deformation can offer critical lead time for risk mitigation and emergency response efforts~\citep{adgeo-14-3-2008}.

While ground-based networks, particularly those relying on Global Navigation Satellite Systems (GNSS), have traditionally been used to monitor deformation~\citep{Poland2024}, 
many volcanoes worldwide remain poorly instrumented or entirely unmonitored \citep{loughlin2015global}. This limitation, coupled with the growing availability of publicly accessible satellite data from missions such as Sentinel, has led to a shift toward satellite-based approaches \cite{spaans2016insar}. Among these, Interferometric Synthetic Aperture Radar (InSAR) has emerged as a powerful tool for global monitoring of surface motion~\citep{hanssen2001radar}, advancing critical applications, including glacier monitoring~\cite{diaconu2025multi}, landslide detection~\cite{yang2024identification}, and land subsidence~\cite{chaussard2014land}.

\begin{figure}[t]
    \centering
    \begin{subfigure}{0.48\linewidth}
        \centering
        \includegraphics[width=\linewidth]{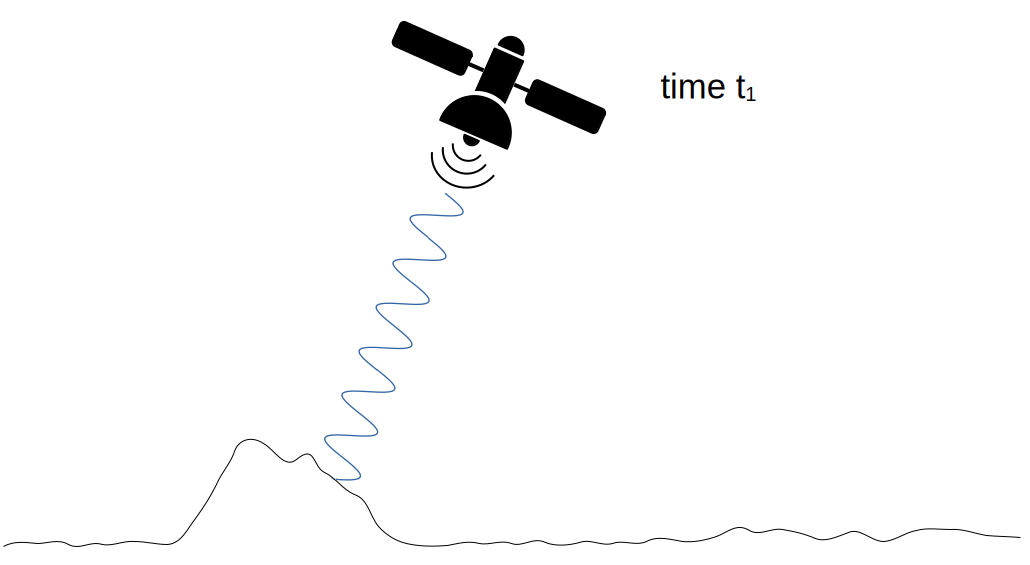}
    \end{subfigure}
    \hfill
    \begin{subfigure}{0.48\linewidth}
        \centering
        \includegraphics[width=\linewidth]{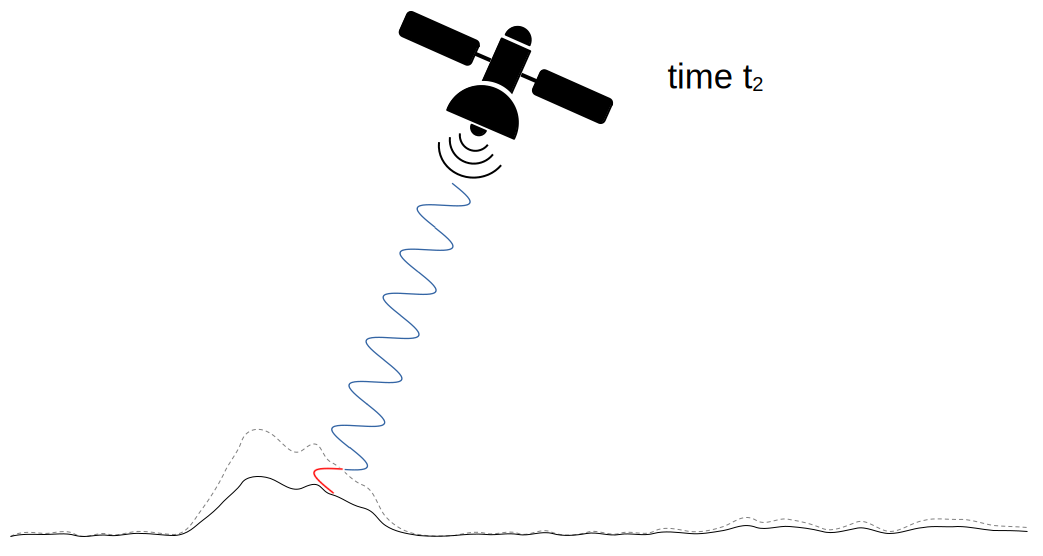}
    \end{subfigure}

    \caption{Schematic description of an InSAR acquisition over a particular location. The satellite passes at a primary acquisition date ($t_{1}$) and a secondary acquisition date ($t_{2}$). In this example, a subsidence event occurs between the two passes, causing the signal to travel a greater distance to reach the ground and return to the satellite at $t_{2}$. By computing the difference between the received signal phases, we obtain the wrapped phase difference map, a cyclic representation of ground deformation where the signal resets every full phase cycle ($2\pi$ radians). Regions with multiple closely-spaced fringes indicate areas of stronger or more concentrated deformation, with the direction of the colour sequence across fringes distinguishing between surface uplift and subsidence.}
    
    \label{fig:insar_overview}
\end{figure}

InSAR estimates surface displacement with millimeter-level precision by analyzing phase differences between two or more SAR acquisitions from the same location at different times. Each acquisition pair consists of a primary date and a secondary date, and the difference in the phase of the received signals between the two dates yields a map of any surface displacement that occurred in the intervening period (see Fig. \ref{fig:insar_overview}). These maps depict displacements as colour fringes, where each fringe represents one full phase cycle ($2\pi$ radians). The number and spacing of fringes, as well as the sequence of colours (ascending or descending in the visible spectrum), provide information on the type of deformation (e.g. subsidence or uplift) and its spatial extent. For Sentinel-1, operating at C-band with a wavelength of approximately 5.6 cm, each full colour cycle corresponds to approximately 2.8 cm of line-of-sight displacement between the ground and the satellite (\cite{braun2021tops}). Since phase measurements can be affected by noise or surface changes, we can assess phase reliability and surface stability by computing the corresponding coherence map. These maps depict the correlation between the phase information of the two SAR images, which quantifies the similarity (coherence) between signals.

Although InSAR is an excellent tool for monitoring ground deformation, its interpretation is highly complex. A major challenge in interpreting InSAR data is distinguishing ground deformation from atmospheric propagation delays.
Lateral variations in ionospheric electron density and tropospheric water vapor concentration can alter the radar signal’s propagation time, introducing phase delays that contaminate the InSAR deformation signal~\citep{zebker1997atmospheric, massonnet1998radar, beauducel2000volcano}. These delays can produce artifacts that mimic real deformation, sometimes manifesting as apparent centimeter-scale ground motion \citep{doin2009corrections}, complicating data interpretation and downstream analysis. 
This is even more prominent in volcanic regions, where complex atmospheric conditions---especially vertical stratification in mountainous terrain---can generate deformation-like patterns.
This increases the risk of false positives, especially over elevated topography such as volcanoes and high-altitude ridges~\citep{PARKER2015102, https://doi.org/10.1029/2011GL049971}.

In our work, we build on the \textit{Hephaestus} 
dataset \citep{bountos_hephaestus_2022} and introduce \textit{Thalia}\footnote{\href{https://en.wikipedia.org/wiki/Thalia_(nymph)}{\textit{Thalia}} (Greek: Th\'aleia, from th\'allein, "to flourish" or "to bloom") was a nymph daughter of \textit{Hephaestus}. \textit{Thalia} symbolizes the renewal of volcanic landscapes after eruptions reflecting the dataset's focus on volcanic activity and its role in data-driven geoscience.}
, enhancing its ground sampling distance and information depth, incorporating additional data sources, and restructuring it to better support time-series analysis. 
\textit{Thalia} introduces a collection of spatiotemporal datacubes that integrate high-resolution InSAR phase difference and coherence products, digital elevation models (DEMs), and lower-resolution atmospheric variables known to confound deformation signals (see \cref{fig:main}). 
Additionally, we include a diverse set of expert annotations that characterize deformation type, intensity, and extent.
Leveraging these improvements, we establish a comprehensive benchmark that demonstrates the ability of \textit{Thalia} to support volcanic activity monitoring as a multimodal, multitemporal classification and semantic segmentation task. 
We report strong and diverse baselines using state-of-the-art models, facilitating fair and reproducible comparison with future methods. We make our dataset and code publicly available at the project's repository \url{https://github.com/Orion-AI-Lab/Thalia}.

\section{Related Work}

Deep learning has been widely used for SAR-based tasks in Earth observation~\citep{9351574,bountos2025kuro,sumbul2021bigearthnet}; however, its application to InSAR remains limited due to the lack of well-curated datasets. This is primarily caused by the high cost of the annotation process, which demands expert knowledge and the scarcity of volcanic activity.


To overcome these challenges and alleviate the need for time-consuming manual annotation, many works leveraged synthetically generated datasets and models pretrained on optical tasks. In particular, one previous study \citep{anantrasirichai_application_2018} relied on major data augmentations and transfer learning from ImageNet \citep{deng2009imagenet}. Building on this, several works focused on synthetically generated InSAR data to train Convolutional Neural Networks (CNNs) for ground deformation detection 
\citep{brengman2021identification,anantrasirichai2019deep, gaddes_simultaneous_2024}.
Similarly, \citet{valade2019towards} utilized synthetic data to train a CNN to predict the associated phase gradients and phase decorrelation mask, which can later be used to detect ground displacement, while \citet{beker2023deep} utilized a synthetic dataset to train CNNs to detect subtle ground deformation from velocity maps.
Another study \citep{bountos2022learning} proposed to train transformer architectures on synthetically generated InSAR using a prototype learning framework,
assigning classes with a nearest-neighbor approach comparing the sample's representation with the class prototypes.

\citet{bountos_self-supervised_2022} diverged from this line of research and proposed the utilization of in-domain self-supervised contrastive learning to create reliable feature extractors without the need for human annotations, emphasizing the performance improvement compared to pretrained weights from optical tasks. In a separate line of work, \citet{popescu2024anomaly} proposed formulating the volcanic ground deformation identification problem as an anomaly detection task using Patch Distribution Modeling \citep{defard2021padim}.  

Given the gradual evolution of volcanic activity, time-series analysis is critical for effective monitoring, with several works exploring this direction.
For example, ~\citet{sun2020automatic} trained a CNN on synthetic data, generated using the Mogi model as the deformation source. They created 20,000 time-series groups, each containing 20 consecutive pairs of unwrapped surface displacement maps. 
Moreover, ~\citet{ansari2021insar} proposed an unsupervised pipeline to cluster similar displacement patterns from InSAR time-series, building on a Long Short-Term Memory (LSTM) autoencoder \citep{hochreiter1997long}, a recurrent deep learning architecture designed to model sequential dependencies, and Hierarchical Density-Based Spatial Clustering of Applications with Noise (HDBSCAN) \citep{campello2013density}.

Finally, \textit{Hephaestus} \citep{bountos_hephaestus_2022} was introduced to address the scarcity of deep learning datasets in the InSAR domain. The dataset provides multi-modal expert annotations and enables the exploration of deep learning approaches for a range of ground deformation analysis tasks, including emerging directions such as InSAR captioning and text-to-InSAR generation. Despite these contributions, \textit{Hephaestus} still presents several limitations, which we discuss in detail in~\cref{subsec:building_on_hephaestus}.

\section{Dataset Construction}
\label{sec:dataset_construction}

\subsection{Building on Hephaestus}
\label{subsec:building_on_hephaestus}
The \textit{Hephaestus} dataset represents a significant step towards advancing \removed{data-driven approaches for volcanic activity monitoring} automated, deep learning-based approaches for volcanic activity monitoring.
Although it offers rich, expert annotations across a global set of volcanoes, its effectiveness in high-precision and time-series geophysical analysis is limited by several factors. First, the spatial resolution of the annotated imagery is relatively coarse, at approximately \(333\,\text{m} \times 333\,\text{m}\) per pixel. Second, the dataset consists of RGB composites of the InSAR products, resulting in loss of physically interpretable pixel values and geolocation information. Finally, the dataset is not designed for spatiotemporal modeling, lacking a machine learning-friendly format.
Recognizing both the promise and the limitations of \textit{Hephaestus}, we take steps to redefine the dataset by addressing its weaknesses and expanding its scope.

\subsection{Thalia}
\label{subsec:minicubes}
\begin{figure}
\resizebox{\textwidth}{!}{
    \centering
    \begin{subfigure}[b]{0.3\textwidth}
        \includegraphics[width=\linewidth]{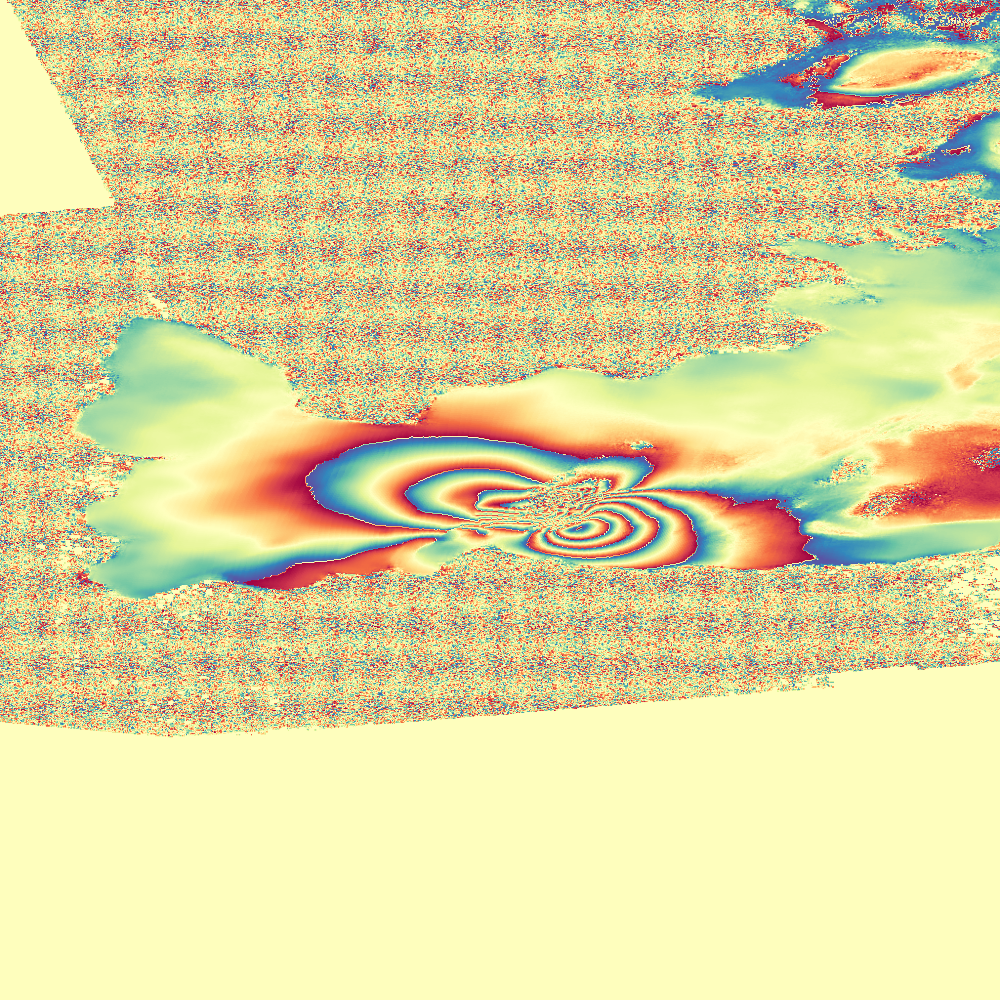}
        \caption{Dyke}
    \end{subfigure}
    \hfill
    \begin{subfigure}[b]{0.3\textwidth}
        \includegraphics[width=\linewidth]{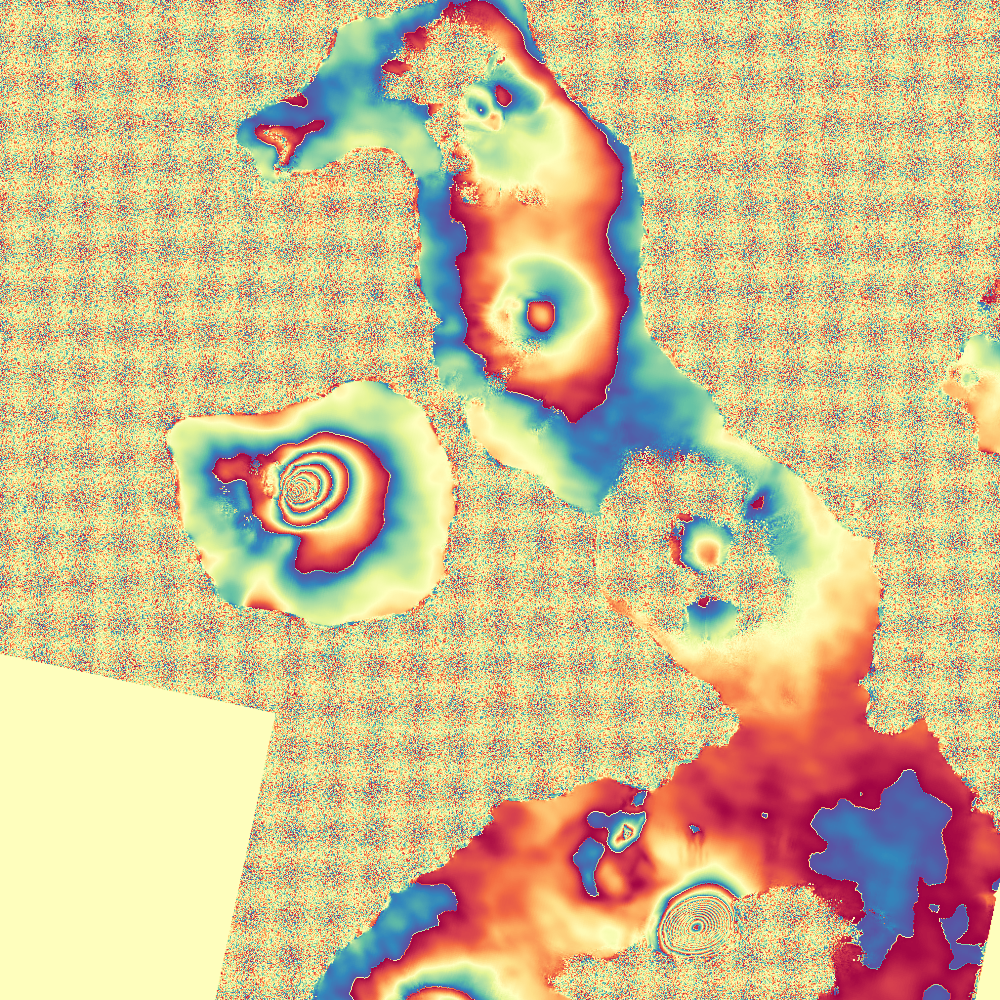}
        \caption{Mogi}
    \end{subfigure}
    \hfill
    \begin{subfigure}[b]{0.3\textwidth}
        \includegraphics[width=\linewidth]{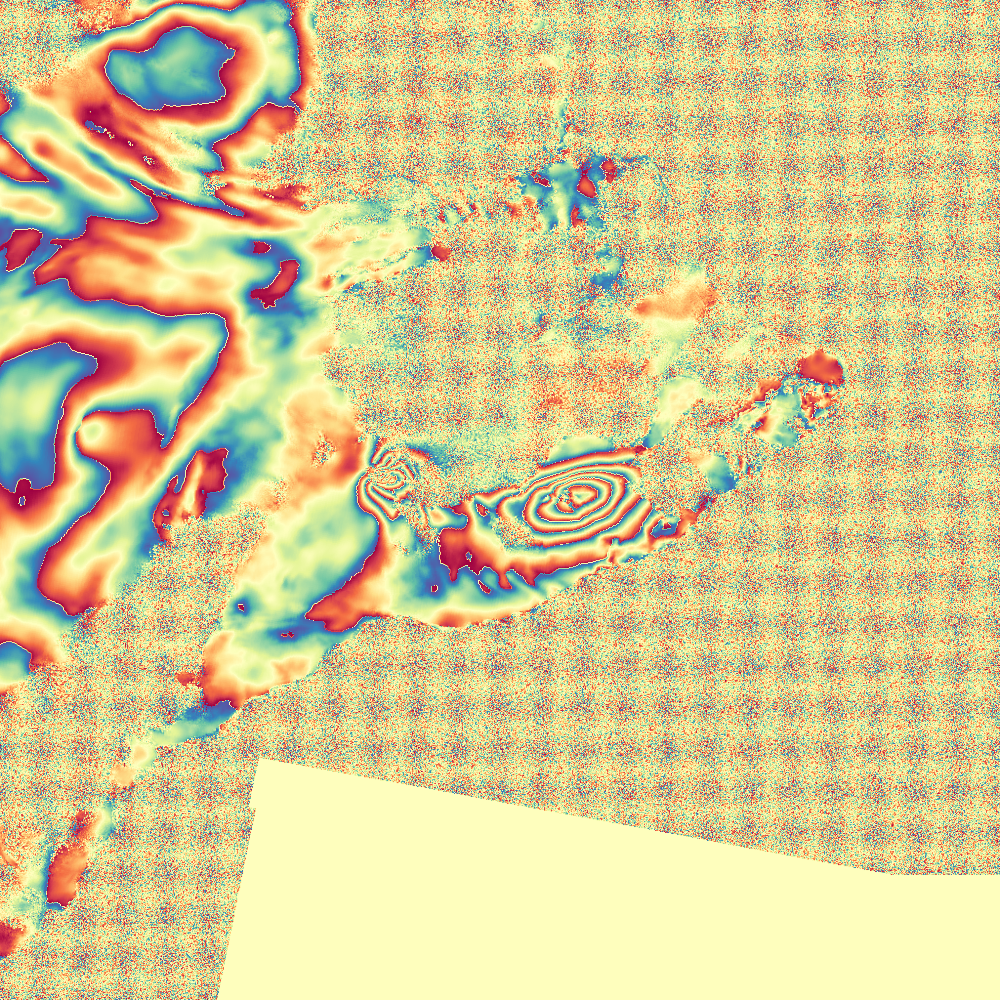}
        \caption{Sill}
    \end{subfigure}
    \hfill
    \begin{subfigure}[b]{0.3\textwidth}
        \includegraphics[width=\linewidth]{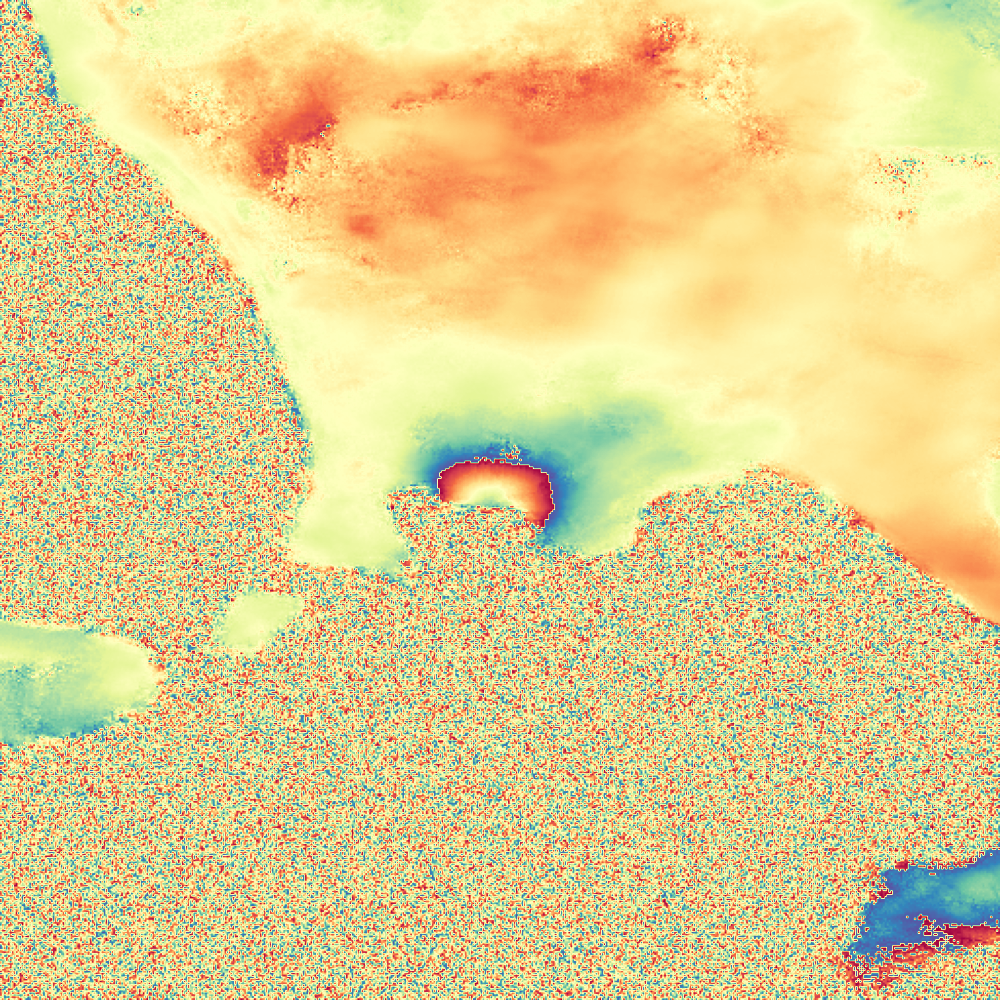}
        \caption{Spheroid}
    \end{subfigure}
    \hfill
    \begin{subfigure}[b]{0.3\textwidth}
        \includegraphics[width=\linewidth]{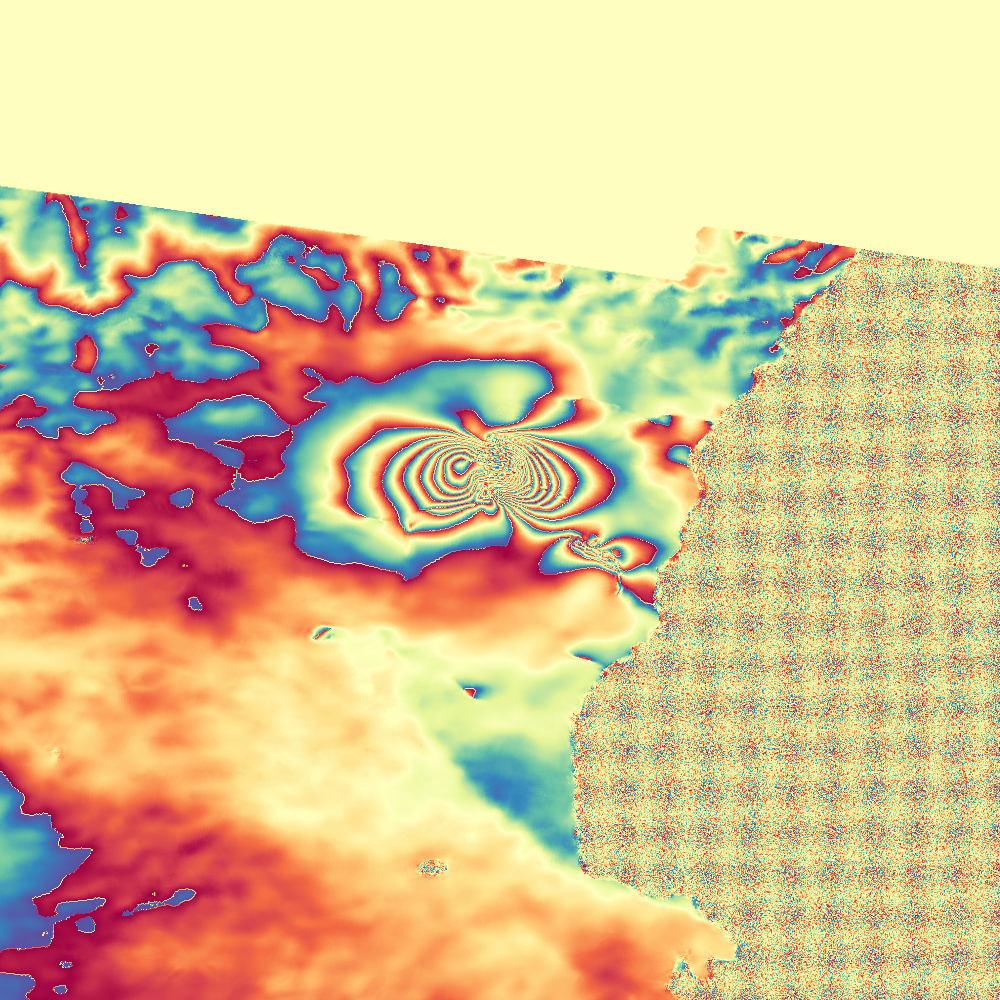}
        \caption{Earthquake}
    \end{subfigure}
    \hfill
    \begin{subfigure}[b]{0.3\textwidth}
        \includegraphics[width=\linewidth]{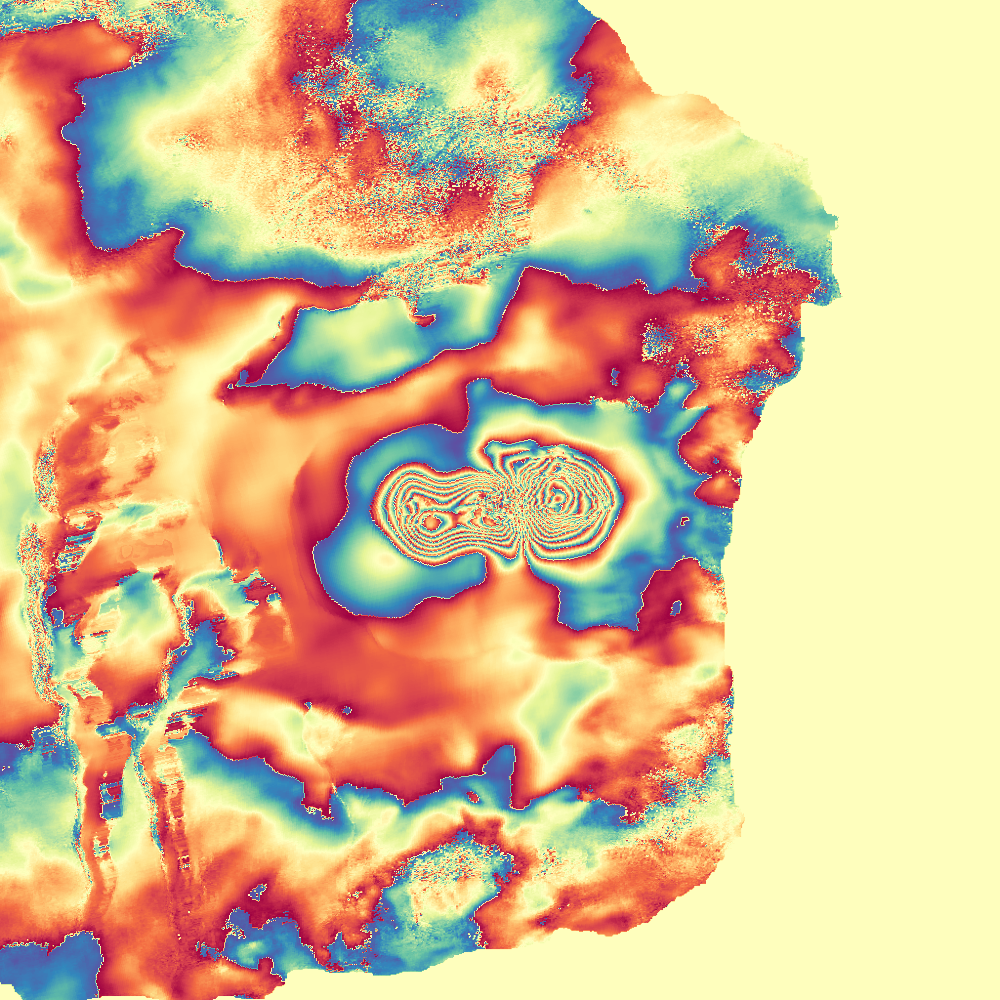}
        \caption{Unidentified}
    \end{subfigure}
}
    
    \caption{Representative examples of ground deformation classes in \textit{Thalia}, including (a) Dyke, (b) Mogi, (c) Sill, (d) Spheroid, (e) Earthquake, and (f) Unidentified patterns. Each class reflects distinct deformation mechanisms relevant to volcanic and tectonic processes.}

    \label{fig:activity_types}
\end{figure}

In this work we introduce \textit{Thalia}, a collection of 38 datacubes covering 44 of the most active volcanoes globally from 2014 to 2021, with a significantly enhanced spatial resolution of approximately \(100\,\text{m} \times 100\,\text{m}\) per pixel, containing a total of 19,942 annotated samples across all datacubes.
The 38 datacubes are derived from COMET-LiCSAR processing frames \citep{lazecky2020licsar}, each covering a fixed geographic region that includes one or more target volcanoes. Frame extents vary in size and may overlap at certain volcanic regions, collectively spanning all 44 sites.
Each datacube \citep{mahecha2020earth} is a structured, multi-dimensional array that co-registers multiple data sources - InSAR products, topographic data, and atmospheric variables - over a fixed geographic region and across time. Unlike storing each modality as independent files, this format preserves spatial and temporal alignment across sources, enabling direct multi-modal and multi-temporal analysis without additional preprocessing. The datacubes are stored in a compressed \texttt{Zarr} format \citep{miles2020zarr}, as structured multi-dimensional arrays optimized for efficient spatiotemporal analysis, with the full dataset totaling 1.7 TB. In the following paragraphs, we describe each component of the \textit{Thalia} dataset, along with important design choices made during its development.

\begin{table*}[]
  \caption{Frequency of each activity type in \textit{Thalia}. Def. and Non-Def. stand for deformation and non-deformation respectively.}
    \label{tab:dataset_label_statistics}
  \centering
  \scriptsize
  \setlength{\tabcolsep}{0.55\tabcolsep}
  \resizebox{\textwidth}{!}{%
  \begin{tabular}{lccccccccccccccccc}
    \toprule
    & \multicolumn{3}{c}{\textbf{Label}} & \multicolumn{6}{c}{\textbf{Activity Type}} & \multicolumn{4}{c}{\textbf{Intensity Level}} & \multicolumn{4}{c}{\textbf{Phase}} \\
    \cmidrule(lr){2-4} \cmidrule(lr){5-10} \cmidrule(lr){11-14} \cmidrule(lr){15-18}
    & Non-Def. & Def. & Earthquake & Sill & Dyke & Mogi & Earthquake & Unidentified & Spheroid & Low & Medium & High & None & Rest & Unrest & Rebound & None \\
    \midrule
    Count & 18089 & 1798 & 55 & 1258 & 527 & 333 & 55 & 50 & 25 & 908 & 533 & 751 & 55 & 18089 & 1664 & 134 & 15 \\
    \bottomrule
  \end{tabular}
  }
\vfill
\end{table*}
\par

\textbf{Dataset Intent.} \textit{Thalia} is designed to support the development of deep learning-based approaches for volcanic activity monitoring, with a focus on identifying characteristic deformation patterns rather than precise displacement quantification. The dataset is built around the wrapped InSAR phase difference, the primary product used by experts for visual interpretation of volcanic deformation. While software exists to unwrap the signal and recover deformation in physical units, this process introduces its own complexities and biases \citep{chen2002phase}. \textit{Thalia} therefore follows expert practice, providing wrapped interferograms alongside rich annotations that reflect how domain experts identify and characterize deformation patterns.

\textbf{InSAR Products.}
The InSAR component lies at the core of \textit{Thalia}, including: a) the wrapped \textit{phase difference}, which captures surface displacement between SAR acquisitions, and b) the \textit{coherence}, which measures the quality of the interferometric signal.
These products are acquired by the COMET-LiCSAR system, which processes Sentinel-1 imagery for global volcano surveillance, in a resolution of approximately \(100\,\text{m} \times 100\,\text{m}\) per pixel. For more information on the InSAR generation and processing pipeline, readers are referred to \citep{lazecky2020licsar}.

\textbf{Topography.}
Stratified atmospheric noise is often correlated with topography. To capture this we include the DEM from LiCSAR, based on the 1 arc-second Shuttle Radar Topography Mission DEM \citep{farr2007shuttle}. This static variable is downsampled for each frame to match the resolution of the InSAR products ($\approx$ \(100\,\text{m} \times 100\,\text{m}\)). 

\textbf{Atmospheric Component.}
A key advancement of \textit{Thalia} is the explicit integration of atmospheric variables known to directly contribute to phase delays in the SAR signal. 
These delays may produce patterns in InSAR imagery that closely resemble actual surface deformation \citep{zebker1997atmospheric, massonnet1998radar, beauducel2000volcano}.
Motivated by established atmospheric correction approaches \citep{yu2018interferometric}, we incorporate atmospheric variables that represent humidity, temperature, and pressure.
Specifically, we include \textit{Total Column Water Vapor}, \textit{Surface Pressure}, and the \textit{Vertical Integral of Temperature}, from the ERA5 single-level reanalysis dataset \citep{hersbach2020era5}, for both primary and secondary SAR acquisition dates. 
We prioritize vertically integrated atmospheric measures, as the impact of atmospheric delays is not confined to specific atmospheric layers. We select the ERA5 data closest in time to each SAR acquisition and resample them to align with the spatial resolution of the InSAR data.

\par
\textbf{Expert Annotations.}
\textit{Thalia} builds upon the manually curated annotations provided by \textit{Hephaestus}, adapting them to the datacube format by converting relevant labels into spatiotemporal masks and carefully addressing differences in spatial resolution and alignment. The available spatiotemporal masks include multi-level information on \textit{ground deformation}, \textit{activity type} (\eg, \texttt{Dyke}, \texttt{Sill}, \texttt{Spheroid}, etc.; see \cref{fig:activity_types}), and \textit{intensity level}  (\eg, \texttt{Low}, \texttt{Medium}, \texttt{High}), while the related volcano’s \textit{phase}  (\eg, \texttt{Rest}, \texttt{Unrest}, \texttt{Rebound}) is represented as a categorical variable (see \cref{tab:dataset_label_statistics}). Additional annotations provide auxiliary information on the presence and type of noise, quality of the samples, annotator confidence and a textual description, offering expert commentary and annotation rationale. Detailed information on all available labels is provided in Supplementary Material (Suppl.) \ref{sec:dataset_description}.

\section{Benchmark}
\label{sec:benchmark}

\begin{table*}[t]
  \centering
  \caption{Summary of class distributions for the benchmark datasets and the leave-3-out evaluation.}
  \label{tab:data_splits}

  \begin{subtable}[t]{0.59\linewidth}
    \centering
    \caption{Main temporal split.}
    \label{tab:temporal_split}
      \resizebox{0.95\linewidth}{!}{
  \begin{tabular}{@{}ccccccc@{}}
    \toprule
    \multirow{2}{*}{Split} & \multirow{2}{*}{Dates} 
    & \multicolumn{2}{c}{Single-Timestep} 
    & \multicolumn{2}{c}{Time-Series} \\
    \cmidrule(lr){3-6}
     & & Positives & Negatives & Positives & Negatives \\
    \midrule
    Training   & Jan 2014 -- May 2019 & 1143 & 8697 & 701 & 2626 \\
    Validation & Jun 2019 -- Dec 2019 & 154  & 2416 & 75  & 728  \\
    Test       & Jan 2020 -- Dec 2021 & 509  & 5992 & 225 & 1776 \\
    \midrule
    Sum        & Jan 2014 -- Dec 2021 & 1806 & 17105 & 1001 & 5130 \\
    \bottomrule
  \end{tabular}
  }
  \end{subtable}
  \hfill
  \begin{subtable}[t]{0.39\linewidth}
    \centering
    \caption{Leave-3-out test set by volcanic region.}
    \label{tab:spatiotemporal_split}
    \resizebox{0.9\linewidth}{!}{
  \begin{tabular}{@{}lcc@{}}
    \toprule
    \multirow{2}{*}{Volcanic Region} 
    & \multicolumn{2}{c}{Single-Timestep} \\
    \cmidrule(lr){2-3}
     & Positives & Negatives \\
    \midrule
    Galapagos & 866 & 318 \\
    East African Rift System & 13 & 712 \\
    South Aegean Volcanic Arc & 0 & 351  \\
    \midrule
    Sum & 879 & 1381 \\
    \bottomrule
  \end{tabular}
  }
  \end{subtable}
\end{table*}

To enable a fair comparison of future methods for InSAR-based volcanic activity detection, we provide the first benchmark on \textit{Thalia}.
This benchmark is designed to serve as a strong baseline across two fundamental tasks: binary ground deformation classification and semantic segmentation. 
To transform our problem into a binary task, we group all sub-classes of ground deformation (\eg, \texttt{Mogi}, \texttt{Sill}, etc.) into one class which we refer to as the positive class. Samples belonging to this class, i.e., interferograms exhibiting ground deformation of any type, are thus referred to as positive samples, while non-deformation samples constitute the negative class.
Below we present the main decisions made for the experimental setup.
Detailed information on the experimental framework is provided in Suppl. \ref{supp:experimental_details}.

\textbf{Data Split.} We apply a temporal data split, separating samples by their InSAR primary acquisition date. The training set includes samples with primary SAR acquisition dates between January 1, 2014, and May 31, 2019, while the validation set covers the period from June 1, 2019, to December 31, 2019. Finally, the test set comprises samples collected between January 1, 2020, and December 31, 2021. This split was carefully designed to preserve spatial diversity across all available frames and to maintain a balanced ratio of positive samples in each set, as shown in~\cref{tab:temporal_split}.

\begin{figure}
    \centering
 %
 \resizebox{0.7\linewidth}{!}{
    \begin{tikzpicture}
        \node[left, rotate=90] at (-3.4, 1.9) {\textbf{Primary Date}};

        \node at (0.8, 3) {\textbf{Secondary Dates}};
        
        \draw[thick] (-3, 2.5) -- (4.5, 2.5);

        \draw[thick, ->] (-3,0.1) -- (-0.5, 0.1);
        \draw[thick, ->] (-3,-0.5) -- (2, -0.5);
        \draw[thick, ->] (-3,-1) -- (4.5, -1);

        \foreach \x in {-3, -0.5,2,4.5} {
            \draw[fill=black] (\x, 2.5) circle (2pt);
        }

        \draw[dashed] (-3, 2.5) -- (-3, -1);
        \draw[dashed] (-0.5, 2.5) -- (-0.5, 0.1);
        \draw[dashed] (2, 2.5) -- (2, -0.5);
        \draw[dashed] (4.5, 2.5) -- (4.5, -1);

        \node at (-1.8, 1.3) {\includegraphics[width=2.2cm]{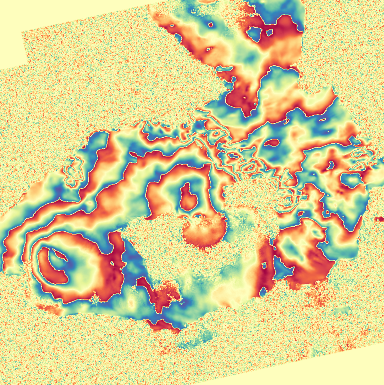}};
        \node at (0.8,0.8) {\includegraphics[width=2.2cm]{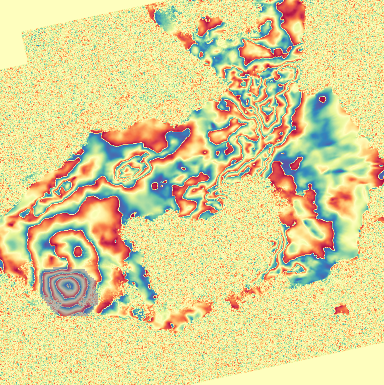}};
        \node at (3.3,0.3) {\includegraphics[width=2.2cm]{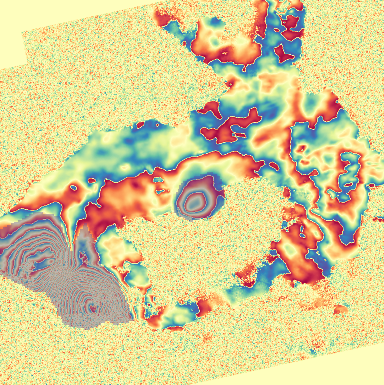}}; 

        \node at (5.5, 0.75) {\includegraphics[height=3.5cm]{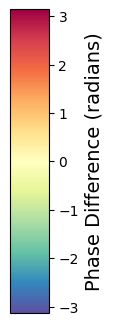}};

    \end{tikzpicture}
    }
 \caption{Schematic illustration of the InSAR time-series construction process. A single primary acquisition date is paired with multiple secondary dates, forming a temporal sequence of interferograms. Each frame shows the wrapped phase difference, with deformation regions highlighted by overlaid masks.}

    \label{fig:time-series_method}
\end{figure}

\textbf{Data Preparation.} 
To maintain a consistent input size, each sample is cropped to \(512 \times 512\) pixels. 
We apply cropping with a bounded random offset from the frame center, constraining positive samples such that at least part of the deformation mask is always included within the cropped area.
We address class imbalance between deformation and non-deformation samples, through data augmentation and undersampling of the negative class, using all available positive samples and an equal number of random negative samples in each epoch. 

\textbf{Constructing InSAR Time-Series.} \label{para:constructing_time-series}Unlike conventional computer vision datasets, constructing meaningful time-series is non-trivial due to the bi-temporal nature of InSAR products, characterized by both primary and secondary acquisition dates. Moreover, the temporal gap between these acquisitions (\ie, the period between the primary and secondary SAR acquisitions that form the InSAR) is not fixed,  
making temporal ordering highly ambiguous \citep{bountos_hephaestus_2022}. In our framework, we define a valid time-series as a sequence of samples where their corresponding InSAR products share the same primary and different secondary acquisition dates. These sequences are ordered chronologically based on the secondary dates, as illustrated in ~\cref{fig:time-series_method}. This formulation exposes models to the evolution of the deformation relative to a fixed reference, allowing them to effectively capture the progression of the deformation patterns. At the same time, varying secondary acquisitions introduces atmospheric noise variations, encouraging the learning of more robust and discriminative features \citep{https://doi.org/10.1029/2001RG000107}.

Following this definition, the number of valid InSAR products per primary SAR acquisition date varies across the dataset. To maintain a consistent input shape for training, \removed{we either select all subsets that match the target sequence length or apply controlled duplication of available products when the desired length cannot be achieved. After examining the distribution of available secondary dates for each primary date, we choose to construct time-series of length three, aiming for a balance between a rich temporal sequence and limited duplications} we fix the time-series length to three, i.e., each time-series consists of three interferograms sharing the same primary acquisition date but paired with three different secondary acquisition dates. When the number of available secondary dates for a given primary date exceeds three, we select all possible subsets of length three. When fewer than three secondary dates are available, we apply controlled duplication of existing products to reach the required time-series length. After examining the distribution of available secondary dates for each primary date, we chose this length of three as a balance between a rich temporal context and limited duplications. For a more detailed explanation of this process we refer the reader to Suppl. \ref{supp:experimental_details}.

We assign a label for each time-series by aggregating across the sequence. 
In particular, for the classification task, a sequence is considered positive if at least one of the products is labeled as showing deformation. For the segmentation task, the target mask is defined as the union of all individual deformation masks across the sequence. This approach ensures that models can leverage temporal information while maintaining a single target.

\textbf{Models.} We employ a diverse set of state-of-the-art models, widely used in Earth observation benchmarks \citep{lacoste2023geo}, 
assessing their capacity for ground deformation classification and segmentation. For the classification task, we include ResNet-50 \citep{he2016deep}, Vision Transformer (ViT) \citep{dosovitskiy2020image}, ConvNeXt \citep{liu2022convnet}, MobileNetV3 \citep{Howard_2019_ICCV}, and EfficientNetV2 \citep{pmlr-v139-tan21a}, all pretrained on ImageNet \citep{deng2009imagenet}. 
For the segmentation task, we use UNet \citep{ronneberger2015u}, DeepLabv3 \citep{chen2017rethinking} and SegFormer \citep{xie2021segformer}, with ResNet-50 backbones pretrained on ImageNet. Additionally, for the time-series setting we include a ConvLSTM \citep{shi2015convolutional} model with a ResNet-50 encoder pretrained on ImageNet for both classification and segmentation tasks, which explicitly models the sequential nature of the input through recurrent processing, in contrast to the channel-stacking approach adopted by the other architectures.

\textbf{Evaluating Input Contributions.} Exploiting the diverse information of \textit{Thalia}, we examine the models' performance across varying configurations to evaluate the importance of each available data source. 
First, we examine the significance of temporal context in detecting volcanic activity on both single-timestep and time-series setups. Second, we assess the impact of the auxiliary atmospheric variables by evaluating the models' performance with and without them. In ~\cref{tab:classification_results,tab:segmentation_results}, we present the classification and segmentation results, respectively, reporting Precision, Recall, F1-score, and Area Under the Receiver Operating Characteristic curve (AUROC) for the classification task,  and Precision, Recall, F1-score, and Intersection over Union (IoU) for the segmentation task. These metrics are chosen specifically to provide a reliable evaluation under the natural class imbalance of the dataset, where deformation samples constitute approximately $10\%$ of the total (see Tab. \ref{tab:dataset_label_statistics}). Formal definitions of all reported metrics are provided in Suppl.~\ref{supp:experimental_details}. 
For all experiments, we report the average performance along with standard deviation over three random seeds.


\begin{table*}
\caption{Deformation classification metrics (mean ± std) for best model configurations between different random seeds. The tables report Precision (Prec), Recall (Rec), F1-score (F1), and Area Under the Receiver Operating Characteristic curve (AUROC) for the deformation class. For all metrics, higher values indicate better performance 
($\uparrow$). The symbols \cmark{} and \xmark{} in the Atm. column denote models trained with and without atmospheric variables as additional input, respectively.} The best value in each column is marked in \textbf{bold}, and the second best is \underline{underlined}, computed 
separately for single-timestep and time-series settings.
\label{tab:classification_results}
\centering
\resizebox{0.7\textwidth}{!}{
\begin{tabular}{@{}lcc|cccc@{}}
\toprule
 & \textbf{Model} & \textbf{Atm.} & \textbf{Prec \added{($\uparrow$)}} & \textbf{Rec \added{($\uparrow$)}} & \textbf{F1 \added{($\uparrow$)}} &\textbf{AUROC \added{($\uparrow$)}} \\
\midrule
\multirow{12}{*}{\textbf{Single-Timestep}} 
 & \multirow{2}{*}{ResNet-50} & \xmark & 83.63 ± 2.94 & 68.5 ± 3.05 & 75.29 ± 2.74 & \textbf{96.99} ± 0.39 \\
 & & \checkmark & 87.53 ± 3.7 & 64.37 ± 1.64 & 74.18 ± 2.32 & \underline{96.26} ± 0.88 \\
 \cmidrule(lr){2-7}
 \cmidrule(lr){2-7}
 & \multirow{2}{*}{MobileNetV3} & \xmark & \textbf{95.03} ± 1.17 & 64.77 ± 0.88 & 77.02 ± 0.37 & 92.06 ± 2.61 \\
 & & \checkmark & 89.56 ± 0.54 & \underline{69.02} ± 1.79 & 78.06 ± 1.13 & 91.99 ± 2.32 \\
 \cmidrule(lr){2-7}
 & \multirow{2}{*}{EfficientNetV2} & \xmark & 76.28 ± 3.74 & 44.66 ± 2.71 & 56.18 ± 1.08 & 74.98 ± 4.13 \\
 & & \checkmark & 82.28 ± 3.55 & 49.44 ± 4.22 & 61.61 ± 3.1 & 83.25 ± 5.02 \\
 \cmidrule(lr){2-7}
  & \multirow{2}{*}{ConvNeXt} & \xmark & 93.04 ± 1.85 & \textbf{69.09} ± 2.01 & \underline{79.25} ± 0.69 & 90.01 ± 3.88 \\
 & & \checkmark & \underline{93.58} ± 0.3 & 68.76 ± 1.89 & \textbf{79.26} ± 1.33 & 90.29 ± 1.75 \\
 \cmidrule(lr){2-7}
  & \multirow{2}{*}{ViT} & \xmark & 85.16 ± 11.21 & 55.8 ± 10.03 & 67.3 ± 10.51 & 87.58 ± 5.02 \\
 & & \checkmark & 90.75 ± 1.51 & 59.27 ± 7.47 & 71.45 ± 5.45 & 88.6 ± 3.0 \\
\midrule

\multirow{14}{*}{\textbf{Time-Series}} 
 & \multirow{2}{*}{ResNet-50} & \xmark & 67.79 ± 2.11 & 60.0 ± 0.36 & 63.64 ± 0.89 & \underline{92.68} ± 1.84 \\
 & & \checkmark & 68.65 ± 0.85 & 59.41 ± 3.27 & 63.66 ± 2.08 & 88.0 ± 2.33 \\
 \cmidrule(lr){2-7}
 \cmidrule(lr){2-7}
 & \multirow{2}{*}{MobileNetV3} & \xmark & 64.08 ± 1.38 & 63.56 ± 3.82 & 63.79 ± 2.54 & 89.39 ± 1.16 \\
 & & \checkmark & 63.51 ± 4.46 & 65.48 ± 5.88 & 64.29 ± 3.71 & 89.29 ± 1.06 \\
 \cmidrule(lr){2-7}
 & \multirow{2}{*}{EfficientNetV2} & \xmark & 68.58 ± 2.94 & 49.04 ± 7.26 & 56.88 ± 5.36 & 82.22 ± 3.5 \\
 & & \checkmark & 64.23 ± 9.07 & 56.0 ± 2.97 & 59.42 ± 4.2 & 82.63 ± 1.54 \\
 \cmidrule(lr){2-7}
  & \multirow{2}{*}{ConvNeXt} & \xmark & 73.19 ± 2.02 & 68.89 ± 15.12 & 70.36 ± 8.63 & 91.65 ± 5.01 \\
 & & \checkmark & 75.88 ± 7.14 & 57.48 ± 2.55 & 65.36 ± 4.3 & 78.24 ± 3.18 \\
 \cmidrule(lr){2-7}
  & \multirow{2}{*}{ViT} & \xmark & \textbf{80.52} ± 5.61 & 53.48 ± 4.62 & 63.92 ± 1.69 & 91.21 ± 3.26 \\
 & & \checkmark & 71.19 ± 2.74 & 61.63 ± 13.4 & 65.54 ± 8.5 & 89.2 ± 3.42 \\
  \cmidrule(lr){2-7}
  & \multirow{2}{*}{\added{ConvLSTM}} & \added{\xmark} & \added{72.89 ± 3.6} & \added{\underline{78.22 ± 3.46}} &  \added{\underline{75.33} ± 1.51} &  \added{92.12 ± 1.45} \\
 & & \added{\checkmark} & \added{\underline{77.01} ± 0.38} &  \added{\textbf{80.89} ± 2.21} & 
  \added{\textbf{78.89} ± 1.09} & \added{\textbf{96.19} ± 0.85} \\

\bottomrule
\end{tabular}
}
\end{table*}

\begin{table*}
\caption{Deformation segmentation metrics (mean ± std) for best model configurations between different random seeds. The tables report Precision (Prec), Recall (Rec), F1-score (F1), and Intersection over Union (IoU) for the deformation class. \added{For all metrics, higher values indicate better performance 
($\uparrow$). The symbols \cmark{} and \xmark{} in the Atm. column denote models trained with and without atmospheric variables as additional input, respectively.} The best value in each column is marked in \textbf{bold}, and the second best is \underline{underlined}\added{, computed 
separately for single-timestep and time-series settings.}}
\label{tab:segmentation_results}
\centering
\resizebox{0.7\textwidth}{!}{
\begin{tabular}{@{}lcc|ccccc@{}}
\toprule
 & \textbf{Model} & \textbf{Atm.} & \textbf{Prec \added{($\uparrow$)}} & \textbf{Rec \added{($\uparrow$)}} & \textbf{F1 \added{($\uparrow$)}} &\textbf{IoU \added{($\uparrow$)}} \\
\midrule
\multirow{6}{*}{\textbf{Single-Timestep}} 
 & \multirow{2}{*}{DeepLabv3}  & \xmark & 81.41 ± 1.42 & \textbf{63.74} ± 0.52 & \textbf{71.49} ± 0.30 & \textbf{55.63} ± 0.37 \\
 & &  \checkmark & 81.64 ± 1.77 & 60.82 ± 2.61 & 69.65 ± 1.48 & 53.46 ± 1.75 \\
 \cmidrule(lr){2-7}
 & \multirow{2}{*}{UNet}  & \xmark & \underline{82.43} ± 0.64 & 61.25 ± 1.27 & \underline{70.27} ± 0.63 & \underline{54.17} ± 0.75 \\
 & & \checkmark & 81.70 ± 0.41 & 53.71 ± 1.89 & 64.80 ± 1.48 & 47.95 ± 1.62 \\
 \cmidrule(lr){2-7}
  & \multirow{2}{*}{SegFormer} & \xmark & 80.87 ± 1.74 & \underline{61.32} ± 1.96 & 69.70 ± 0.70 & 53.50 ± 0.82 \\
 & &  \checkmark & \textbf{83.13 ± 2.30} & 55.71 ± 0.98 & 66.68 ± 0.22 & 50.01 ± 0.25 \\
\midrule
\multirow{8}{*}{\textbf{Time-Series}} 
 & \multirow{2}{*}{DeepLabv3} & \xmark & 75.68 ± 2.15 & 54.66 ± 1.54 & 63.42 ± 0.25 & 46.44 ± 0.27 \\
 & &  \checkmark & 74.64 ± 2.48 & 46.67 ± 2.15 & 57.39 ± 1.84 & 40.27 ± 1.82 \\
 \cmidrule(lr){2-7}
 & \multirow{2}{*}{UNet} & \xmark & 74.57 ± 3.10 & 58.57 ± 2.39 & 65.50 ± 0.35 & 48.7 ± 0.39 \\
 & &  \checkmark & 70.28 ± 4.56 & 44.18 ± 1.64 & 54.19 ± 2.03 & 37.2 ± 1.92 \\
 \cmidrule(lr){2-7}
  & \multirow{2}{*}{SegFormer} & \xmark & \textbf{79.22} ± 0.55 & 57.87 ± 1.46 & 66.87 ± 0.77 & 50.23 ± 0.87 \\
 & & \checkmark & 77.14 ± 0.10 & 47.88 ± 2.73 & 59.05 ± 2.10 & 41.92 ± 2.12 \\
  \cmidrule(lr){2-7}
  & \multirow{2}{*}{\added{ConvLSTM}} & \added{\xmark} &  \added{\underline{78.81} ± 4.76} &  \added{\textbf{71.93} ± 0.73} &  \added{\textbf{75.16} ± 2.52} &  \added{\textbf{60.27} ± 3.18} \\
 & & \added{\checkmark} & \added{77.55 ± 4.17} & \added{\underline{66.50} ± 2.37} & \added{\underline{71.48} ± 1.47} & \added{\underline{55.64} ± 1.80}\\
\bottomrule
\end{tabular}
}
\end{table*}

\subsection{Generalization Across Unseen Volcanoes}
\label{subsec:generalization_on_unseen_volc}
Our benchmark is designed to simulate an operational setting in which models are expected to monitor a known, pre-defined set of active volcanoes. 
As a complementary investigation---separate from the main benchmark---we consider an additional split to examine the models’ ability to generalize to \textit{previously unseen volcanoes}, thereby assessing their robustness to distribution shifts induced by spatiotemporal alterations, changes in topography, and variations in environmental conditions. However, constructing a representative spatial split is not straightforward as each volcano may exhibit unique activity frequency that manifests differently in InSAR data influenced by its geomorphology and structural characteristics. In this investigation, we define a leave-3-out split, in which three volcanic regions---(a) the Galápagos Islands, (b) the East African Rift System, and (c) the South Aegean Volcanic Arc---are reserved for testing, and are removed from the training and validation sets. These test regions were selected to represent three qualitatively distinct activity regimes: (a) high-frequency deformation activity, (b) low-frequency deformation activity, and (c) no recorded deformation activity. All remaining volcanoes are used for training (January 1, 2014 – December 31, 2019) and validation (January 1, 2020 – December 31, 2021). ~\cref{tab:spatiotemporal_split} presents 
the class distributions for the reserved test regions. 

\begin{table*}[ht!]
\caption{Deformation classification and segmentation metrics (mean ± std) between different random seeds on the leave-3-out split for the models that performed best in the temporal split. 
Each subtable reports Precision (Prec), Recall (Rec), F1-score (F1), and a) AUROC or b) IoU for the deformation class. For all metrics, higher values indicate better performance ($\uparrow$). The symbols \cmark{} and \xmark{} in the Atm. column denote models trained with and without atmospheric variables as additional input, respectively.} \removed{The best value in each column is marked in \textbf{bold}, and the second best is \underline{underlined}.}
\label{tab:classification_segmentation_spatial}
\centering
\renewcommand{\arraystretch}{0.9} 
\setlength{\tabcolsep}{3pt}       

\begin{subtable}{0.8\textwidth} 
\centering
\caption{Classification metrics.}
\label{tab:classification_spatial}
\scriptsize
\begin{tabular}{@{}lcc|cccc@{}}
\toprule
 & \textbf{Model} & \textbf{Atm.} & \textbf{Prec \added{($\uparrow$)}} & \textbf{Rec \added{($\uparrow$)}} & \textbf{F1 \added{($\uparrow$)}} & \textbf{AUROC \added{($\uparrow$)}} \\
\midrule
\multirow{2}{*}{\textbf{Single-Timestep}} 
  & \multirow{2}{*}{ConvNeXt} & \xmark & 59.97 ± 0.23 & 86.63 ± 1.97 & 70.87 ± 0.81 & 83.95 ± 0.96 \\
  & & \checkmark & 60.31 ± 1.89 & 90.83 ± 0.95 & 72.46 ± 1.05 & 85.8 ± 1.74 \\
\midrule
\multirow{2}{*}{\textbf{\added{Time-Series}}} 
  & \multirow{2}{*}{\added{ConvLSTM}} & \added{\xmark} & \added{ 72.6 ± 1.79} & \added{93.6 ± 0.5} & \added{81.77 ± 1.24} & \added{ 89.44 ± 0.89} \\
  & & \added{\checkmark} & \added{ 76.08 ± 3.67} & \added{ 87.61 ± 4.41 } & \added{ 81.37 ± 3.28} & \added{ 89.37 ± 2.61} \\
\bottomrule
\end{tabular}
\end{subtable}

\vspace{3mm} 

\begin{subtable}{0.8\textwidth}
\centering
\caption{Segmentation metrics.}
\label{tab:segmentation_spatial}
\scriptsize
\begin{tabular}{@{}lcc|cccc@{}}
\toprule
 & \textbf{Model} & \textbf{Atm.} & \textbf{Prec \added{($\uparrow$)}} & \textbf{Rec \added{($\uparrow$)}} & \textbf{F1 \added{($\uparrow$)}} & \textbf{IoU \added{($\uparrow$)}} \\
\midrule
\multirow{2}{*}{\textbf{Single-Timestep}} 
  & \multirow{2}{*}{DeepLabv3} & \xmark & 35.32 ± 2.56 & 88.07 ± 1.25 & 50.35 ± 2.48 & 33.68 ± 2.20 \\
  & & \checkmark & 24.65 ± 3.47 & 90.82 ± 2.00 & 38.62 ± 4.14 & 24.01 ± 3.20 \\
\midrule
\multirow{2}{*}{\textbf{\added{Time-Series}}} 
  & \multirow{2}{*}{\added{ConvLSTM}} & \added{\xmark} & \added{26.54 ± 2.3} & \added{93.04 ± 1.18} & \added{41.23 ± 2.7} & \added{26.01 ± 2.14} \\
  & & \added{\checkmark} & \added{36.06 ± 5.71} & \added{92.62 ±  1.33} & \added{51.59 ± 5.87} & \added{34.97 ± 5.24} \\
\bottomrule
\end{tabular}
\end{subtable}

\end{table*}

\section{Discussion}
\label{sec:discussion}

\textbf{Overall performance. }Examining the performance of the classification baselines in \cref{tab:classification_results}, we observe strong discriminative capability reaching up to $\approx79\%$ in F1-Score. This performance declines in the segmentation task, reaching up to $\approx75\%$ (\cref{tab:segmentation_results}). This is not a surprising behavior, as the exact delineation of ground deformation is often non-trivial even for experts. Even after discerning true ground deformation from atmospheric contributions, defining the extent of such fringes is ambiguous, especially in regions with high incoherence (see \cref{subsec:model_insights}) or strong relief. Such noise is inherent to the data itself, making annotation and thereby accurate prediction challenging \cite{kondylatos2025probabilistic}. Many works have sought to improve segmentation capabilities under such conditions 
\cite{acuna2019devil,yu2018simultaneous}. Our benchmark establishes a strong reference point for future methods that aim to address these complexities.

\textbf{Impact of temporal dimension. }

While the theoretical advantages of using time-series data to capture volcanic activity and account for atmospheric delays are well established \citep{https://doi.org/10.1029/2001RG000107},
direct performance comparison between single-timestep and time-series inputs is not straightforward in our setting.
Although both approaches aim to detect the same underlying geophysical phenomena, they operate on different data subsets due to the stricter requirements for constructing valid time-series (see \cref{para:constructing_time-series}), which reduces the overall dataset size and alters the ratio of positive to negative samples (see \cref{tab:temporal_split}). 
More importantly, the task formulation shifts: single-timestep models operate on individual images, predicting class labels for classification or deformation masks for segmentation. In contrast, time-series models process multiple acquisitions jointly, indicating, for classification, the presence of at least one positive observation, and for segmentation, the union of deformation patterns capturing the total extent of the affected area. As such, while performance trends are informative, comparison in absolute metrics between the two setups should be interpreted with these structural differences in mind.
Through our experiments we observe comparable performance in the classification task and superior performance in the segmentation task when using time-series with a recurrent model architecture such as ConvLSTM \citep{shi2015convolutional}. This is expected given that the model benefits from richer temporal context on the extent of deformation, and the temporal aggregation of target labels mitigates the effect of atmospheric noise. On the other hand, we observe a notable decline in performance in both tasks when relying on channel stacking, highlighting that the effective exploitation of temporal context requires architectures specifically designed for sequential modeling.

\par
\textbf{Impact of atmospheric variables. }
Atmospheric variables contain valuable information regarding the atmospheric conditions present in both SAR acquisitions, but are available at a substantially coarser spatial resolution than the InSAR data. This mismatch has different implications for each task: while they offer contextual information on the likelihood of atmospheric signal contribution, they cannot directly help in precise delineation.  
\Cref{tab:classification_results} demonstrates that the majority of the classification models (8/10) benefit from the inclusion of atmospheric information (\textit{atmospheric-setting}), exhibiting performance gains against models trained only on InSAR modalities and DEM (\textit{base-setting})
in both single-timestep and time-series inputs.  
In contrast, all segmentation models fail to effectively utilize atmospheric variables, often leading to deteriorated performance compared to the \textit{base-setting}. 

We hypothesize that this discrepancy reflects limitations of early fusion to combine such fundamentally distinct data sources---high-resolution deformation signals and coarse atmospheric context---for precise ground deformation delineation, rather than the atmospheric data lacking valuable information.
We investigate this hypothesis through weight analysis and ablation studies (see Suppl. \ref{supp:fusion_inv}), which indicate that while atmospheric variables are actively incorporated by the models' early fusion layers, their incorporation compromises the models' ability to extract fine-scale deformation features and degrades the quality of the \textit{base-setting} modalities representation, as evidenced by degraded unimodal performance compared to models trained exclusively in the base-setting.
These findings suggest that early fusion 
may introduce a detrimental trade-off for segmentation as the fusion layer's limited capacity must be shared between high-resolution, deformation-specific features, and coarse atmospheric context. Alternative architectures that preserve separate representations for multi-resolution modalities (\eg, late fusion or cross-attention) may better leverage complementary information without this capacity bottleneck. Our benchmark provides a testbed to assess such models in both \textit{atmospheric} and \textit{base-settings}.

\textbf{Generalization on Unseen Volcanoes.}
As discussed in \cref{subsec:generalization_on_unseen_volc}, 
assessing generalization to unseen volcanoes is inherently challenging, as no single spatial split can represent the diversity of volcanic activity.
Nevertheless, our experiments 
indicate strong generalization capability in the classification task (\cref{tab:classification_spatial}), where we observe a moderate F1-score reduction of $\approx8\%$. For the segmentation task (\cref{tab:segmentation_spatial}), the performance drop is more pronounced, $\approx20$–$25\%$ in both single-timestep and time-series input, primarily due to oversensitivity in detecting deformation signals, as reflected by lower precision values. We observe the same trend regarding the inclusion of the atmospheric variables as in the temporal split, \ie, improvement in the classification task and performance drop in segmentation.

\subsection{Qualitative Results and Model Behavior}
\label{subsec:model_insights}

\begin{figure}[H]
    \centering
    \begin{subfigure}{0.48\linewidth}
        \centering
        \includegraphics[width=\linewidth]{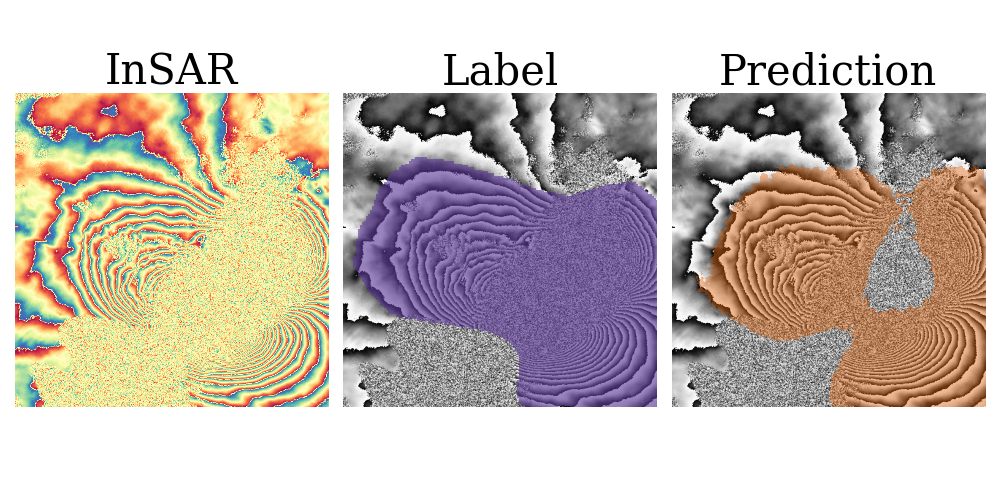}
    \end{subfigure}
    \hfill
    \begin{subfigure}{0.48\linewidth}
        \centering
        \includegraphics[width=\linewidth]{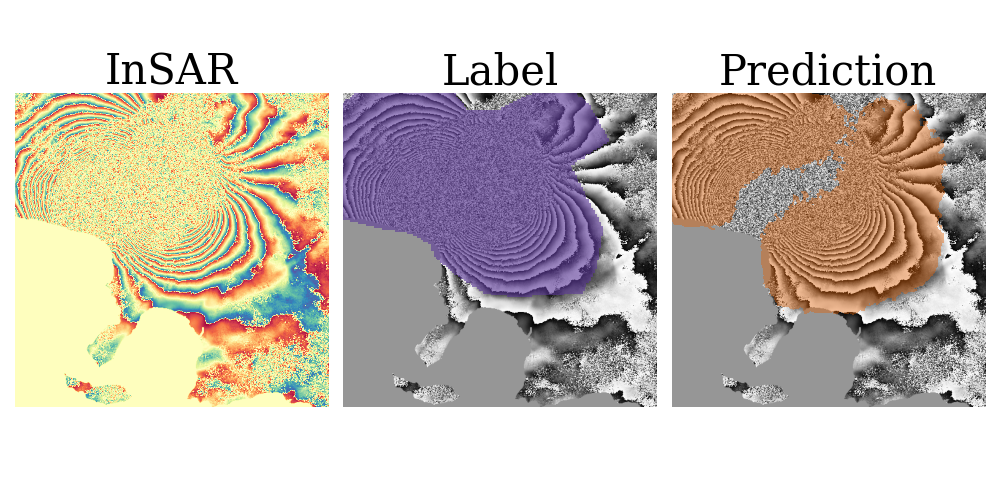}
    \end{subfigure}

    \caption{Qualitative examples of true positive predictions over Taal, Philippines, illustrating the ambiguity in defining deformation extent. Each group of images shows the InSAR phase difference (left), as well as the corresponding ground truth in purple (middle) and model prediction in orange (right) over a grayscale rendering of the InSAR product. Results are from the best-performing DeepLabv3 model with atmospheric input.}
    
    \label{fig:model_ambiguity_examples}
\end{figure}

In this section, we present indicative examples that provide insights into the performance of models beyond quantitative metrics, especially given the inherent ambiguities and complexities of the data.

\par
\textbf{Ambiguity in Segmentation Masks.}
 Delineating the exact boundaries of deformation is often challenging, especially in regions of high incoherence and/or of strong atmospheric signals. Such noise is inherent to the data itself, making the annotation and, thereby, accurate prediction challenging. In \cref{fig:model_ambiguity_examples}, we provide two representative examples of qualitatively good model predictions that do not perfectly align with the annotator's estimation, illustrating the aforementioned ambiguity. In this case, model predictions accurately identify the observable deformation patterns to some extent, but exclude the noisy areas. The human annotator, however, considers these noisy regions part of the event, even if they do not exhibit deformation fringes. 

\par
\begin{figure}[t]
    \centering
    \begin{tabular}{>{\raggedleft}m{0cm} m{13.5cm}}
        (a) &
        \includegraphics[width=0.88\linewidth]{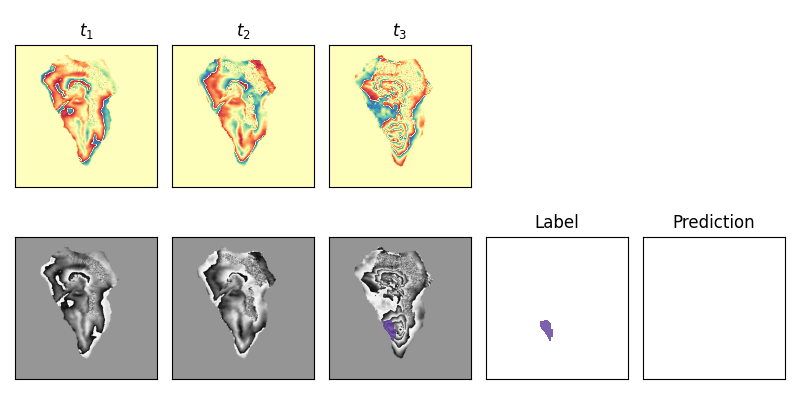} \\
        (b) &
        \includegraphics[width=0.88\linewidth]{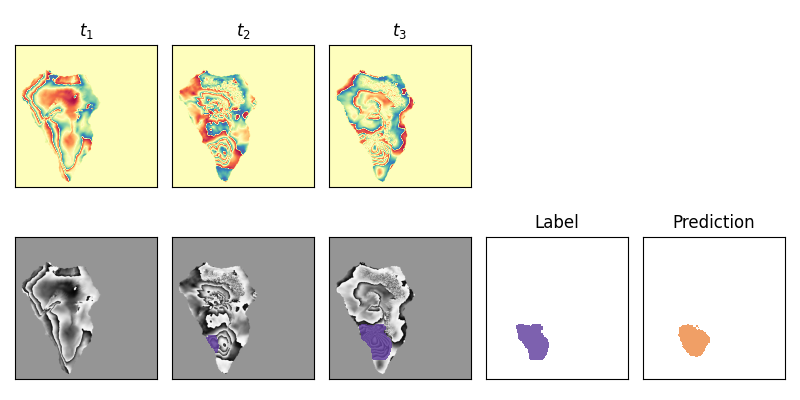} \\
    \end{tabular}
    \caption{Time-series examples over La Palma, Tenerife, illustrating the influence of temporal context on model predictions. The upper row in each subfigure shows InSAR products across timesteps, and the bottom row shows ground truth masks \removed{(\textcolor[RGB]{145,91,96}{red})} \added{(purple)} overlaid on InSAR (grayscale). The final \removed{frame combines all ground truth masks (\textcolor[RGB]{145,91,96}{red}) and model predictions (\textcolor[RGB]{97,131,107}{green})} \added{two frames display the union of all ground truth masks (purple) and model predictions (orange) respectively}. Results are from the best-performing SegFormer model with atmospheric input. Subfigure (a) shows a false negative case with late-onset deformation, while (b) depicts a successful detection when deformation persists over multiple timesteps.}
    \label{fig:ts_seg_examples}
\end{figure}

\textbf{Time-Series Predictions.}
~\Cref{fig:ts_seg_examples} illustrates important insights concerning model predictions on time-series. Although both presented time-series focus on the same volcanic region, variations in the spatial extent or deformation onset across time-steps can affect detection ability.  In the first sequence, subtle deformation patterns are only visible in the final time-step, which the model fails to detect. In contrast, the second sequence displays a progressively intensifying deformation signal starting from the second time-step, providing stronger temporal cues that allow the model to identify and segment the affected area correctly. This example highlights how both the timing and strength of deformation signals across the time-series could play an important role in the model's ability to identify deformation.

\subsection{\removed{Limitations} \added{Inherent Domain Challenges and Dataset Limitations}}
\label{subsec:limitations}

Despite extensive efforts in the annotation process, which incorporates validation from both internal and external sources, InSAR annotation remains inherently challenging. The labels used for training and evaluation are not free from noise, reflecting the complexities involved in detecting volcanic activity. As discussed in \cref{subsec:model_insights}, the subtlety and variability of volcanic deformation patterns often lead to ambiguities in interpretation, making accurate and consistent annotation difficult, especially in regions with low signal coherence. This limitation is compounded by the fact that some volcanic events are subtle or evolve over extended periods, which can further complicate the identification and classification of deformation signals in the data.

Additionally, the temporal scale and nature of volcanic activity introduce significant challenges. Some activity episodes are subtle and unfold over many years, while others are abrupt and short-lived. The frequency and expression of these events vary widely between volcanoes; some may remain inactive for extended periods before suddenly exhibiting signs of activity. Consequently, certain volcanoes may not show any positive samples during the dataset's timeframe, limiting the model’s exposure to their activity patterns. As a result, models trained under such constraints may struggle to generalize effectively in operational settings, particularly when tasked with detecting activity at volcanoes with sparse or no prior positive observations. While our additional leave-3-out experiment (see \cref{subsec:generalization_on_unseen_volc}) provides important insights to that direction, it represents only a subset of possible spatial and temporal variability.

\section{Conclusion}
\label{sec:conclusion}
\textit{Thalia} represents a \removed{significant advancement in data-driven} \added{strong foundation in deep-learning based} volcanic activity monitoring.
By integrating high-resolution InSAR phase and coherence products, DEM, atmospheric information, and expert annotations into structured spatiotemporal datacubes the dataset provides a rich foundation for machine learning research in this domain.
We demonstrated \textit{Thalia's} ability to support volcanic activity monitoring, providing a comprehensive benchmark for both volcanic activity detection and delineation, highlighting important challenges and promising research directions on the utilization of multi-modal and multi-temporal data.


\textit{Thalia} is, to the best of our knowledge, the first large-scale machine learning ready dataset to incorporate such diverse information, enabling interdisciplinary collaboration between the machine learning and Earth Science communities.

\section*{Acknowledgements}
LiCSAR contains modified Copernicus Sentinel data [2014-2021] analysed by the Centre for the Observation and Modelling of Earthquakes, Volcanoes and Tectonics (COMET). LiCSAR uses JASMIN, the UK’s collaborative data analysis environment (\url{http://jasmin.ac.uk}).

\section*{Funding}
This work has received funding from the project ThinkingEarth (grant agreement No 101130544) and from the project MeDiTwin (grant agreement No 101159723) of the European Union’s Horizon Europe research and innovation programme.

\section*{Author Contributions}
\textbf{Nikolas Papadopoulos} (Corresponding Author): Conceptualization (Equal), Data curation (Lead), Investigation (Lead), Methodology (Equal), Software (Lead), Validation (Lead), Visualization (Equal), Writing -- original draft (Equal), Writing -- review \& editing (Equal).

\textbf{Nikolaos Ioannis Bountos}: Conceptualization (Equal), Data curation (Supporting), Investigation (Supporting), Methodology (Equal), Software (Supporting), Validation (Supporting), Visualization (Supporting), Writing -- original draft (Equal), Writing -- review \& editing (Equal).

\textbf{Maria Sdraka}: Data curation (Supporting), Investigation (Supporting),
Methodology (Supporting), Software (Supporting), Validation (Supporting),
Visualization (Equal), Writing -- original draft (Supporting),
Writing -- review \& editing (Supporting).

\textbf{Andreas Karavias}: Conceptualization (Supporting), Data curation (Supporting),
Validation (Supporting), Writing -- original draft (Supporting).

\textbf{Gustau Camps-Valls}: Supervision (Supporting), Writing -- original draft (Supporting),
Writing -- review \& editing (Supporting).

\textbf{Ioannis Papoutsis}: Conceptualization (Equal), Funding acquisition (Lead),
Supervision (Lead), Validation (Supporting), Writing -- original draft (Supporting), Writing -- review \& editing (Supporting).


\section*{Data availability}
The dataset and code are publicly available at the project's repository \texttt{\url{https://github.com/Orion-AI-Lab/Thalia}}.
Our code is released under the MIT \footnote{\href{https://opensource.org/license/MIT}{https://opensource.org/license/MIT}} and data under the CC-BY license \footnote{\href{https://creativecommons.org/licenses/by/4.0/}{https://creativecommons.org/licenses/by/4.0/}}.

\appendix

\section{Detailed Dataset Description}
\label{sec:dataset_description}

This section details the annotation schema used in \textit{Thalia}. Building upon the expert labels from the original \textit{Hephaestus} dataset \citep{bountos_hephaestus_2022}, annotations were adapted to a spatiotemporal format compatible with the datacube structure. A comprehensive overview of the dataset variables is presented in \cref{tab:dataset_overview}, and examples of different annotation types are illustrated in \cref{subsec:examples}. Finally, we provide a detailed outline of the annotation process in \cref{subsec:annotation}.

\subsection{Variables}
\label{subsec:data_vars}

\begin{table*}[!ht]
  \caption{Overview of the variables included in \textit{Thalia}}
  \label{tab:dataset_overview}
  \centering
  \scriptsize
  \begin{tabular}{p{5cm} p{7cm}}
    \toprule
    \textbf{Variable} & \textbf{Description} \\
    \midrule

    \multicolumn{2}{l}{\textbf{InSAR Products}} \\
    Phase Difference & 	Difference in SAR signal phase, indicating surface displacement.\\
    Coherence & The reliability of the interferometric phase measurement.\\

    \midrule
    \multicolumn{2}{l}{\textbf{Topography}} \\
    DEM & Digital Elevation Model: a representation of Earth's surface topography.\\

    \midrule
    \multicolumn{2}{l}{\textbf{Atmospheric Variables}} \\
    Total Column Water Vapor & Water vapor from the surface to the top of the atmosphere. \\
    Surface Pressure & Atmospheric pressure at Earth's surface.\\
    Vertical Integral of Temperature & Mass-weighted temperature integral from surface to top of the atmosphere.\\

    \midrule
    \multicolumn{2}{l}{\textbf{Annotations}} \\
        Deformation Mask & High level mask for deformation presence \{\texttt{Non-Deformation, Volcanic, Earthquake}\}. \\
        Activity Type Mask & Mask identifying the activity type: \{{\texttt{Sill, Dyke, Mogi, Spheroid, Earthquake, Unidentified}}\}. \\
        Intensity Level Mask & Mask identifying the intensity level of the activity \{\texttt{None, Low, Medium, High}\}.\\
        Phase & Phase of the activity: \{\texttt{Rest, Unrest, Rebound, None}\}.\\
        Atmospheric Fringes & Identifies Types of Atmospheric Noise. \\
        Glacier Fringes & Identifies deformation patterns from glacier melting.\\
        Orbital Fringes & Identifies phase ramps due to orbital errors.\\
        Corrupted & Flag for corrupted data. \\
        No Info &  Identifies low-coherence interferograms, where meaningful interpretation is not possible.\\
        Low Coherence & Identifies samples that are characterized from interferometric signal decorrelation.\\
        Is Crowd & Identifies whether multiple deformation masks exist.\\
        Caption & Expert text description of the interferogram and annotation rationale.\\
        Confidence & Value indicating the annotator's confidence [\texttt{0}, \texttt{1}].\\

    \midrule
    \multicolumn{2}{l}{\textbf{Metadata}} \\
    Unique Id & Unique identifier for each sample.\\
        Valid Date Pair & Boolean flag for primary/secondary dates with existing InSAR products.\\
    
    \bottomrule
  \end{tabular}
\end{table*}

\textbf{Activity.} 
Activity-related segmentation masks lie at the core of the dataset’s annotation schema detailing the presence of deformation, its geophysical source and its intensity.  

A high-level \textit{Deformation Mask}, delineates \texttt{Volcanic} and \texttt{Earthquake} induced deformation patterns.

The \textit{Activity Type Mask} captures more detailed geophysical context, distinguishing among several common deformation source models: \texttt{Mogi} \citep{kiyoo1958relations}, \texttt{Dyke} \citep{okada1985surface, sigmundsson2010intrusion}, \texttt{Sill} \citep{fialko2001deformation}, and \texttt{Spheroid} \citep{yang1988deformation}, as well as deformation attributed to \texttt{Earthquake} events or labeled as \texttt{Unidentified} when no clear model can be observed.

The \textit{Intensity Level Mask} categorizes the strength of deformation signals based on the number of visible fringes in the interferogram: \texttt{Low} (1 fringe), \texttt{Medium} (2–3 fringes), and \texttt{High} (more than three fringes). For earthquake-related events, the intensity is marked as \texttt{None}, reflecting their distinct deformation characteristics.

Additionally, we include a \textit{Phase} categorical variable that captures the state of the volcano, \ie\texttt{Rest} (no sign of volcanic activity), \texttt{Unrest} (indicating uplift), or \texttt{Rebound} (indicating subsidence). Again, for Earthquake events, we set the phase to \texttt{None}.

\textbf{Noise.}
A separate set of annotation variables aims to capture specific signal characteristics and noise patterns. These include: \textit{glacier fringes}, when observed fringes result from glacier melting (\texttt{1} if present, \texttt{0} otherwise); \textit{orbital fringes}, for phase ramps caused by satellite orbital errors; and \textit{atmospheric fringes}, which take four values: \texttt{type 0} (no atmospheric impact), \texttt{type 1} (vertical stratification correlated with topography due to changes in the troposphere’s refractive index), \texttt{type 2} (turbulent mixing and vapors caused by liquid or solid particles in the atmosphere), and \texttt{type 3} (a combination of \texttt{type 1} and \texttt{type 2}). The \textit{low coherence} variable is set to \texttt{1} when interferometric signal decorrelation dominates the image. \textit{No info} is used when coherence is so low throughout the interferogram that meaningful interpretation is not possible.

\textbf{InSAR Processing Errors.} 
Automated InSAR generation pipelines may, occasionally, result in faulty products. To identify corrupted InSAR and facilitate the automatic detection of such instances in future applications, we include a set of annotation variables that denote technical faults. 
 The first is the binary variable \textit{corrupted}, which identifies interferograms that are entirely unusable due to being problematic. The second is \textit{processing error}, which distinguishes between specific InSAR processing errors: \texttt{type 1} refers to debursting errors during the synchronization of bursts from one or more sub-swaths; \texttt{type 2} indicates Sentinel-1 sub-swath merging errors, which appear as visible discontinuities, while \texttt{type 0} denotes interferograms free of such processing issues.

 \textbf{Meta-Information.}
 \removed{The \textit{confidence} score is a continuous value in the range \texttt{[0, 1]} that reflects the annotator’s confidence regarding the deformation classification.} The \textit{is crowd} variable is set to \texttt{0} when at most one local fringe pattern is present and \texttt{1} when two or more such patterns appear within the same interferogram. Furthermore, the \textit{caption} field contains a text description providing expert commentary, interpretation, rationale, or relevant contextual notes for the InSAR phase difference product. Finally, the \textit{unique id} serves as a unique identifier for each sample, and the \textit{valid date pair} flag indicates whether the combination of primary and secondary dates corresponds to an existing InSAR product, intended for use in the multi-dimensional array functionality.

\textbf{\added{Annotator Confidence.}}
\added{The \textit{confidence} score takes values in the range $[0, 1]$, providing a per-sample estimate of annotator certainty. Annotations with low confidence were discussed among the group of annotators to reach a consensus. All positive activity annotations were additionally cross-validated against external sources, including peer-reviewed publications, reliable news sources, and the COMET Volcano Deformation Portal, while annotations that remained below $0.5$ confidence were discarded as unreliable.}

\begin{figure}[!]
    \centering
    \includegraphics[width=0.8\linewidth]{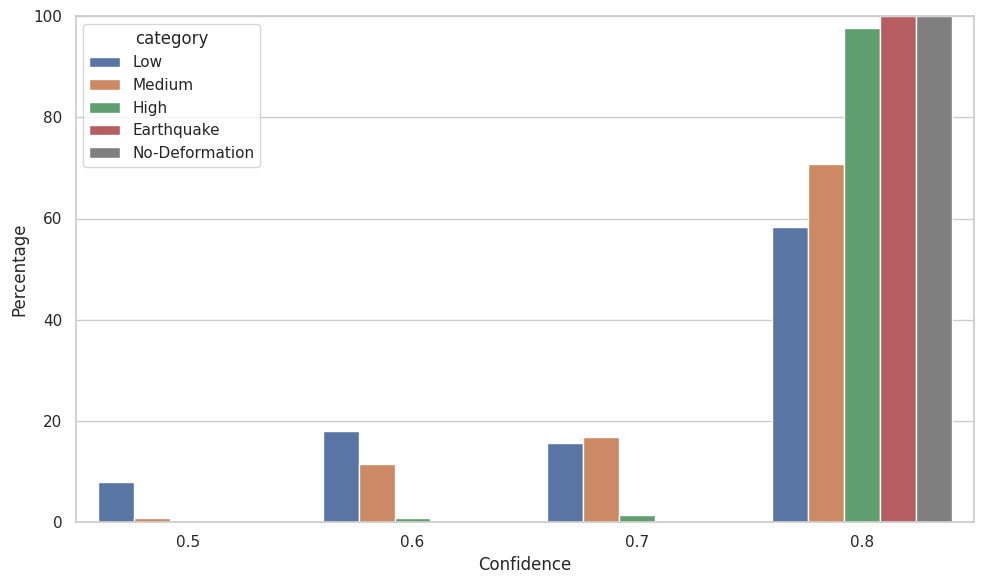}
\caption{\added{Distribution of annotator confidence scores across intensity level
categories. Bars within each category sum to $100\%$. No-Deformation samples 
are annotated with high-confidence, 
reflecting the annotation protocol in which a negative label was assigned only 
when the annotator was certain of the absence of deformation. Among positive 
cases, catalog-confirmed events such as Earthquakes and High-intensity 
deformation are predominantly annotated with high-confidence, while the 
proportion of lower-confidence annotations increases for Medium- and 
Low-intensity events, where deformation patterns are subtler and more 
susceptible to atmospheric contamination.}}
    \label{fig:confidence_intensity}
\end{figure}

\added{The annotation protocol was designed to prioritize recall of subtle deformation. Ambiguous cases where subtle deformation could not be ruled out were retained as low-confidence positives rather than assigned a negative label, while negative labels were assigned only when the annotator was confident of the absence of deformation. This protocol is illustrated in Fig.~\ref{fig:confidence_intensity}, with catalog-confirmed events such as earthquakes and high-intensity deformation predominantly annotated at high confidence levels, whereas the proportion of lower-confidence annotations increases for medium- and low-intensity events, where deformation patterns are subtler and more susceptible to atmospheric contamination.}

\added{We note that even for well-established events, fully certain annotations are rare in practice, as the confidence score captures uncertainty across all annotation aspects — including deformation extent, atmospheric and glacier fringes, and text captions — rather than deformation presence alone.}

\subsection{Visualizing Annotation Diversity}
\label{subsec:examples}
In this section, we present selected examples of different annotation variables, illustrating the diversity and richness of the annotation variables in \textit{Thalia}.

\par
\begin{figure*}[htbp]
  \centering
  \begin{subfigure}[b]{0.32\textwidth}
    \centering
    \includegraphics[width=\textwidth, height=5cm, keepaspectratio=false]{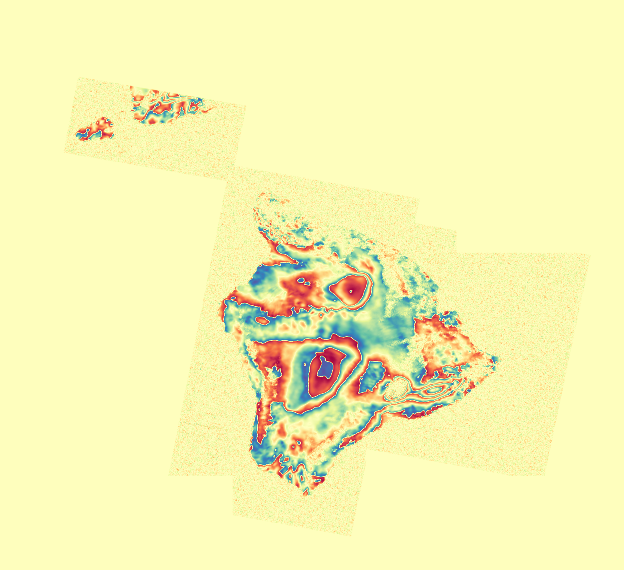}
    \caption{}
    \label{fig:dyke}
  \end{subfigure}
  \hfill
  \begin{subfigure}[b]{0.32\textwidth}
    \centering
    \includegraphics[width=\textwidth, height=5cm, keepaspectratio=false]{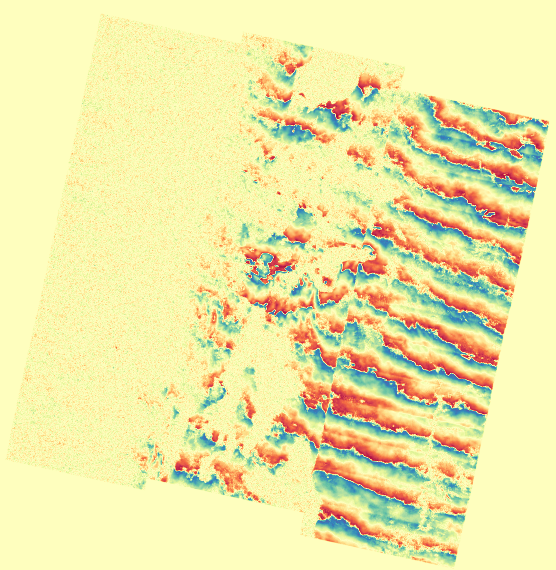}
    \caption{}
    \label{fig:turbulent}
  \end{subfigure}
  \hfill
  \begin{subfigure}[b]{0.32\textwidth}
    \centering
    \includegraphics[width=\textwidth, height=5cm, keepaspectratio=false]{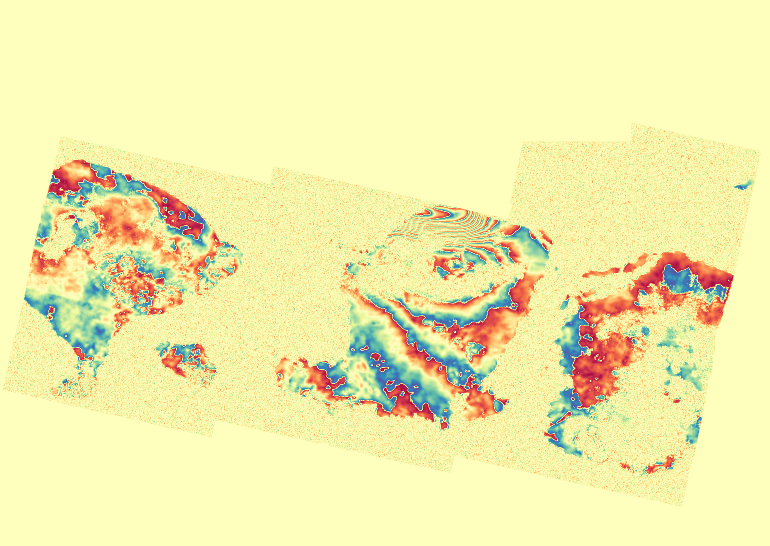}
    \caption{}
    \label{fig:earthquake}
  \end{subfigure}

  \scriptsize
  \begin{flushleft}
(a) ``Vertical stratification can be detected in the high altitude areas. Turbulent mixing effect can also be detected on the right, central, and top-left sides. Two deformation patterns can be detected on the right side. A dyke-type of high intensity on the leftmost and a sill-type of medium intensity on the rightmost."

\par (b) ``Turbulent mixing effect or wave-like patterns caused by liquid and solid particles of the atmosphere can be detected around the area. No deformation activity can be detected. Orbital fringes detected. Difficulties in extracting information."

\par (c) ``Noise can be detected on the bottom-right area. Turbulent mixing effect can be detected on the wider left and central sides of the region. Vertical stratification can also be detected on the central-top side of the region. An earthquake deformation pattern can be detected on the top-central side of the region."
  \end{flushleft}
  \caption{Textual annotations highlighting volcanic and atmospheric phenomena in InSAR imagery from the \textit{Thalia} dataset.}
  \label{fig:caption_examples}
\end{figure*}

\textbf{Text Captions.}
In ~\cref{fig:caption_examples}, we showcase InSAR imagery featuring various apparent deformation patterns and atmospheric phenomena, accompanied by expert textual captions that highlight the complexity and variability captured in the dataset. Each caption provides a thorough description of the location and type of all underlying phenomena, as well as a reference to any interpretation challenges that may be present.

\begin{figure*}[htbp]
    \centering
    \begin{subfigure}[t]{\textwidth}
        \centering
        \includegraphics[width=0.7\textwidth, height=6.5cm, keepaspectratio=false]{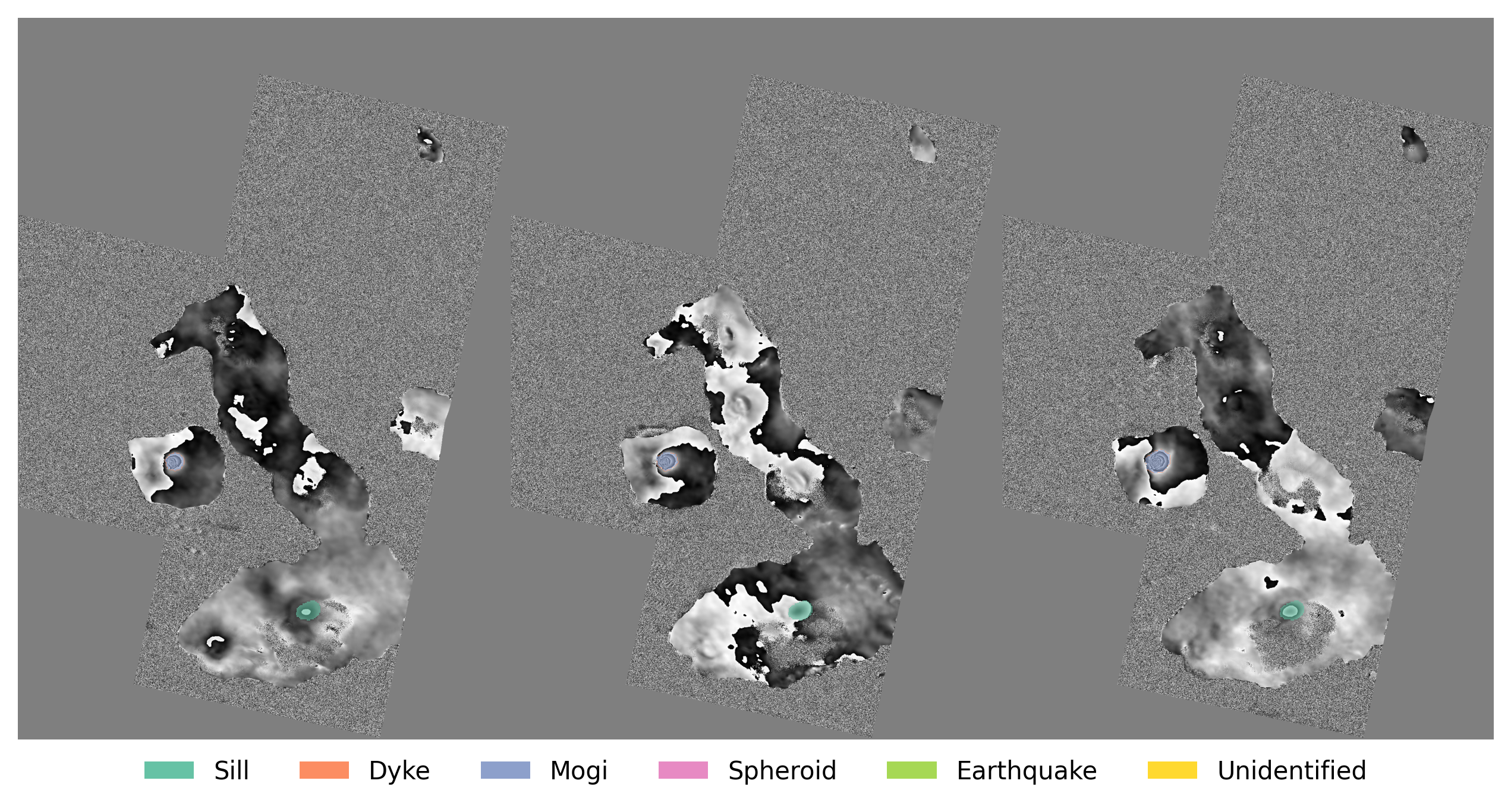}
        \caption{Overlay of \textit{activity type} masks on a time-series of InSAR phase difference products.}
        \label{fig:activity_masks}
    \end{subfigure}

    \vspace{1em} 

    \begin{subfigure}[t]{\textwidth}
        \centering
        \includegraphics[width=0.7\textwidth, height=4cm, keepaspectratio=false]{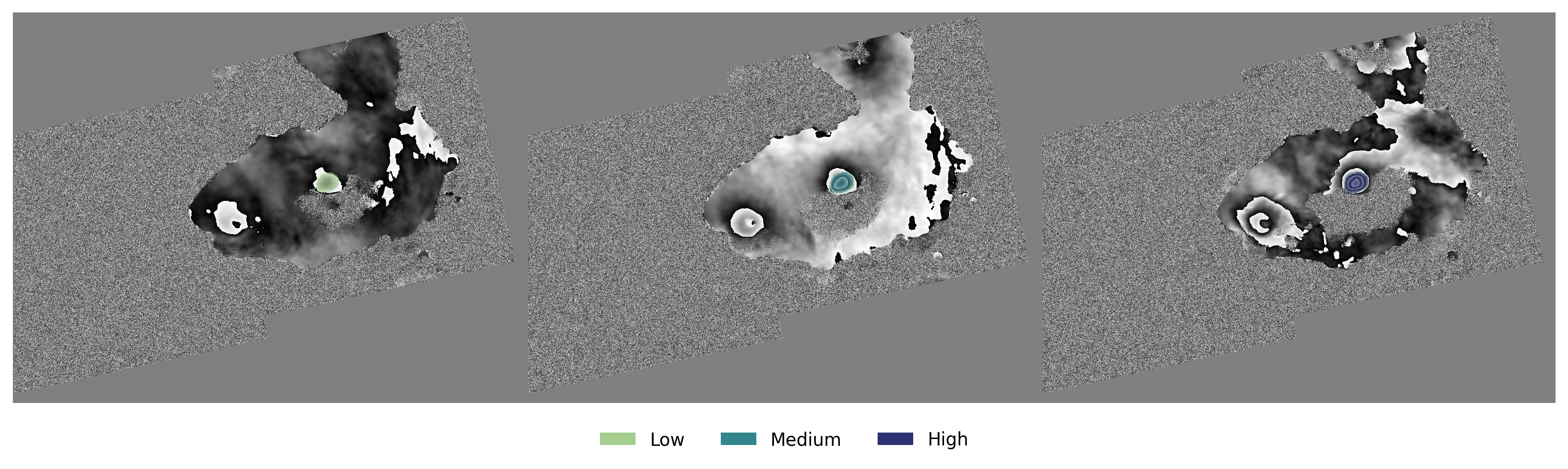}
        \caption{Overlay of \textit{intensity level} masks on a time-series of InSAR phase difference products.}
        \label{fig:intensity_masks}
    \end{subfigure}

    \caption{Examples of annotated InSAR products with overlayed annotated masks indicating: (a) distinct geophysical activity types and (b) activity intensity levels.}
    \label{fig:mask_examples}
\end{figure*}

\textbf{Activity and Intensity Masks.}
In ~\cref{fig:mask_examples}, we present different time series of InSAR products overlaid with expert-provided \textit{activity type} and \textit{intensity level} masks. These examples illustrate the temporal evolution of deformation signals and how distinct geophysical phenomena are annotated both categorically and in terms of intensity, emphasizing the structured labeling within \textit{Thalia}. The InSAR products are shown in grayscale for visual clarity.

\subsection{Annotation Process}
\label{subsec:annotation}

A team of InSAR experts carried out the annotation process through photo-interpretation on the available wrapped interferograms. Each image was optically inspected over the locations of volcanoes, as well as the surrounding areas, to evaluate the quality of the interferograms and identify the impact of atmospheric delays and potential artifacts. Each positive annotation was cross-validated via external sources (publications, reliable news sources) as well as the COMET Volcano Deformation Portal \footnote{\url{https://comet.nerc.ac.uk/comet-volcano-portal/}}, to support the evaluation of whether the observed displacement was due to actual volcanic activity or atmospheric delay effects. For further details regarding the annotation process, readers are referred to the original \textit{Hephaestus} dataset \citep{bountos_hephaestus_2022}.

\section{Extended Experimental Details}
\label{supp:experimental_details}

In this section, we provide additional details on the experimental configurations used in the provided benchmark.



\added{
\textbf{Evaluation Metrics.}
We report the following metrics for both classification and segmentation tasks, 
all computed with respect to the deformation class. $TP$, $TN$, $FP$, and $FN$ 
denote true positives, true negatives, false positives, and false negatives, respectively, 
computed at the sample level for classification and at the pixel level for 
segmentation. For all metrics, higher values indicate better performance 
($\uparrow$).}

\begin{itemize}
    \item \added{\textbf{Precision} ($\uparrow$): The fraction of predicted deformation cases that are correct:}
    \added{
    \begin{equation}
        \text{Precision} = \frac{TP}{TP + FP}
    \end{equation}}

    \item \added{\textbf{Recall} ($\uparrow$): The fraction of true deformation 
    cases that are successfully retrieved:}
    \added{
    \begin{equation}
        \text{Recall} = \frac{TP}{TP + FN}
    \end{equation}}

    \item \added{\textbf{F1-score} ($\uparrow$): The harmonic mean of Precision and 
    Recall, providing a single balanced measure of detection performance:}
    \added{
    \begin{equation}
        \text{F1} = \frac{2 \cdot \text{Precision} \cdot \text{Recall}}
        {\text{Precision} + \text{Recall}}
    \end{equation}}

    \item \added{\textbf{AUROC} ($\uparrow$): The Area Under the Receiver Operating Characteristic curve, a plot visualizing the True Positive Rate ($TP/(TP + FN)$) against the False Positive Rate ($FP/(FP + TN)$) of a binary classifier at various decision thresholds. In our experiments, AUROC measures the model's ability to discriminate between deformation and non-deformation samples. A value of 1.0 indicates perfect discrimination; 0.5 corresponds to random chance. Reported for the classification task only.}

    \item \added{\textbf{IoU} ($\uparrow$): The Intersection over Union (or Jaccard Index), measuring 
    the spatial overlap between the predicted and ground truth deformation masks:}
    \added{
    \begin{equation}
        \text{IoU} = \frac{TP}{TP + FP + FN}
    \end{equation}}
    \added{Reported for the segmentation task only.}
\end{itemize}

\begin{figure*}[ht!]
    \centering
    \includegraphics[width=\textwidth]{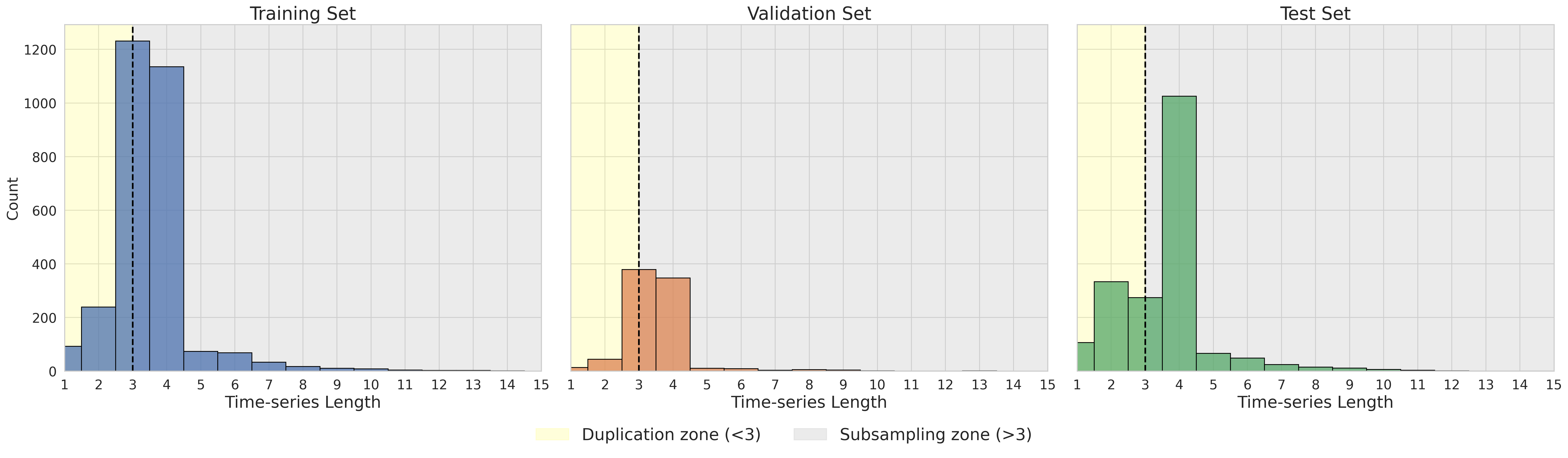}
    \caption{Distribution of time-series lengths across training, validation, and test sets.}
    \label{fig:timeseries_length_distribution}
\end{figure*}

\begin{table*}[ht!]
  \caption{Overview of classification architectures}
  \label{tab:arch_class_overview}
  \centering
  \scriptsize
  \begin{tabular}{p{3cm} p{2cm} p{2cm} p{2cm} p{2cm}}
    \toprule
     & \multicolumn{2}{l}{\hspace{0.5cm} \textbf{\# Trainable parameters (M)}} & \multicolumn{2}{l}{\hspace{0.5cm} \textbf{\# Average runtime (hours)}} \\
    \textbf{Model architecture} & \textbf{Single-image} & \textbf{Time-series} & \textbf{Single-image} & \textbf{Time-series} \\
    \midrule 
    ResNet-50 & 23.5 & 23.6 & 2.2& 2.7\\
    MobileNetV3 (Large) & 4.2 & 4.2 & 1.7& 2.0\\
    EfficientNetV2 (Small) & 20.2 & 20.2 & 2.6 & 2.7\\
    ConvNeXt (Base) & 87.6 & 87.6 & 6.4 & 5.2\\
    ViT (Small) & 22.3 & 24.0 & 2.5 & 3.2\\
    \bottomrule
  \end{tabular}
\end{table*}

\begin{table*}[ht!]
  \caption{Overview of segmentation architectures}
  \label{tab:arch_seg_overview}
  \centering
  \scriptsize
  \begin{tabular}{p{2cm} p{1.5cm} p{1.5cm} p{1.5cm} p{1.5cm} p{1.5cm}}
    \toprule
    & & \multicolumn{2}{l}{\hspace{0.15cm} \textbf{\# Trainable parameters (M)}} & \multicolumn{2}{l}{\hspace{0.15cm} \textbf{\# Average runtime (hours)}} \\
    \textbf{Model architecture} & \textbf{Backbone} & \textbf{Single-image} & \textbf{Time-series} & \textbf{Single-image} & \textbf{Time-series} \\
    \midrule 
    DeepLabV3 & ResNet-50& 26.7 & 26.8 & 2.3 & 2.9\\
    UNet & ResNet-50 & 32.5 & 32.6 & 2.3 & 3.0 \\
    SegFormer & ResNet-50& 24.9 & 24.9 & 1.6 & 3.7 \\
    \bottomrule
  \end{tabular}
\end{table*}

\textbf{Time-series Construction.}
We construct the time-series by first grouping samples based on their primary SAR acquisition date. The distribution of the length of the resulting time-series can be seen in Fig. \ref{fig:timeseries_length_distribution}. Based on this, we fix the time-series length $k$ to 3, to ensure input consistency. For primary SAR acquisitions that produce time-series with length $n$ fewer than 3, we randomly duplicate available interferograms to reach the required time-series length. Conversely, if they contain fewer than 3 we extract all possible sub-sequences of the predefined length, resulting in a number of samples given by the binomial formula in ~\cref{eq:binomial_k3}.

\begin{align}
\binom{n}{k} &= \frac{n!}{k!(n-k)!} \\[10pt]
\overset{k=3}{\Longrightarrow} \binom{n}{3} &= \frac{n!}{3!(n-3)!} \label{eq:binomial_k3}
\end{align}

\begin{figure}[h]
    \centering
    \includegraphics[width=0.8\linewidth]{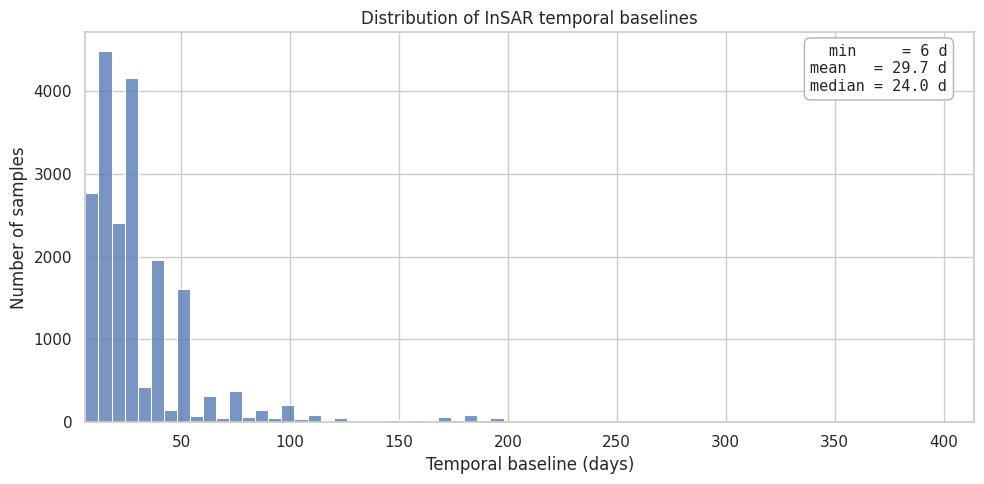}
    \caption{\added{Distribution of temporal baselines (in days) between the primary and secondary SAR
    acquisition dates across all InSAR products in \textit{Thalia}.}}
    \label{fig:temporal_baseline}
\end{figure}

\added{
\textbf{Temporal Baseline.} Each InSAR product in \textit{Thalia} is formed from a pair of SAR
acquisitions: a primary (reference) date and a secondary date. We refer to the time elapsed between
these two acquisitions as the \textit{temporal baseline} of the interferogram. The temporal baseline
is not fixed across the dataset, as it depends on the availability of Sentinel-1 passes over each
COMET-LiCSAR processing frame; it therefore varies from a few days to several weeks. This
quantity is distinct from the time-series length discussed above: the temporal baseline characterizes
the span of an \emph{individual} interferogram, whereas the time-series length refers to the number of
interferograms sharing a common primary date. Figure~\ref{fig:temporal_baseline} shows the
distribution of temporal baselines across all InSAR products in \textit{Thalia}. The distribution is
strongly right-skewed, with a minimum of 6 days, a median of 24 days, and a mean of 29.7 days,
reflecting the predominance of short revisit pairs together with a tail of longer-baseline products.
Because the temporal baseline directly influences the magnitude and evolution of the observed phase
difference, it constitutes an important source of variability that future work could exploit.
}

\paragraph{Training Setup.} We conducted all experiments on a single GPU (NVIDIA GeForce RTX 3090 Ti). 
We trained all models for 90 epochs, using the AdamW optimizer \citep{loshchilov2017decoupled} with a fixed learning rate of $10^{-5}$ and a weight decay set to $10^{-2}$. For classification, we found that cross-entropy loss consistently delivered the best performance, except for ViT that displayed more stable convergence with focal loss. For segmentation, focal loss was more effective, likely due to its robustness to class imbalance in pixel-wise annotations \citep{lin2017focal}. To mitigate the effect of class imbalance, we employed an undersampling strategy during training. For each epoch, all available positive samples were included, along with a randomly selected subset of negative samples of equal size. All models were trained with a random set of data augmentations including gaussian blur, random resize crop, horizontal and vertical flips and random rotations. Tables \ref{tab:arch_class_overview} and \ref{tab:arch_seg_overview} report the number of trainable parameters for all models along with the average runtime of each experiment for both classification and segmentation tasks respectively. 



\section{Investigation of Early Fusion on Segmentation Performance}
\label{supp:fusion_inv}

Although atmospheric variables provide useful contextual information, as reflected by the improved performance of classification models (see Tab. 2 in Sec. 5 of the main paper), their substantially coarser spatial resolution compared to InSAR products can hinder their effective use in segmentation tasks. This is further evidenced by the drop in segmentation performance, suggesting that segmentation models struggle to leverage atmospheric inputs and often perform worse than the \textit{base-setting}. We hypothesize that this discrepancy arises from limitations of early fusion when combining fundamentally different data sources—high-resolution deformation signals and coarse atmospheric context—for precise ground deformation delineation, rather than from a lack of informative content in the atmospheric variables.

\begin{figure*}[htbp]
  \centering
  
  \begin{subfigure}{0.3\textwidth}
    \includegraphics[width=\linewidth]{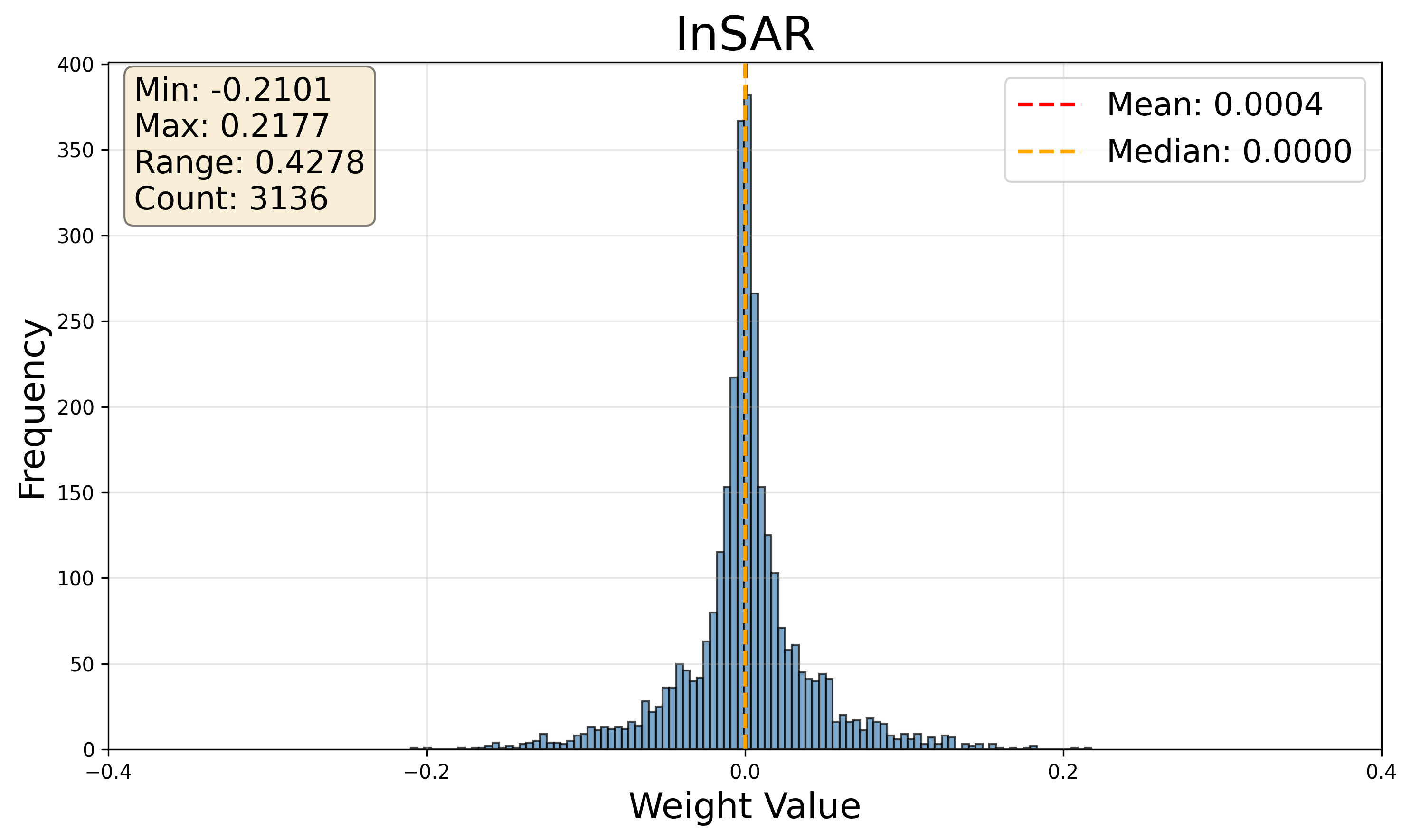}
    \caption{}
  \end{subfigure}\hfill
  \begin{subfigure}{0.3\textwidth}
    \includegraphics[width=\linewidth]{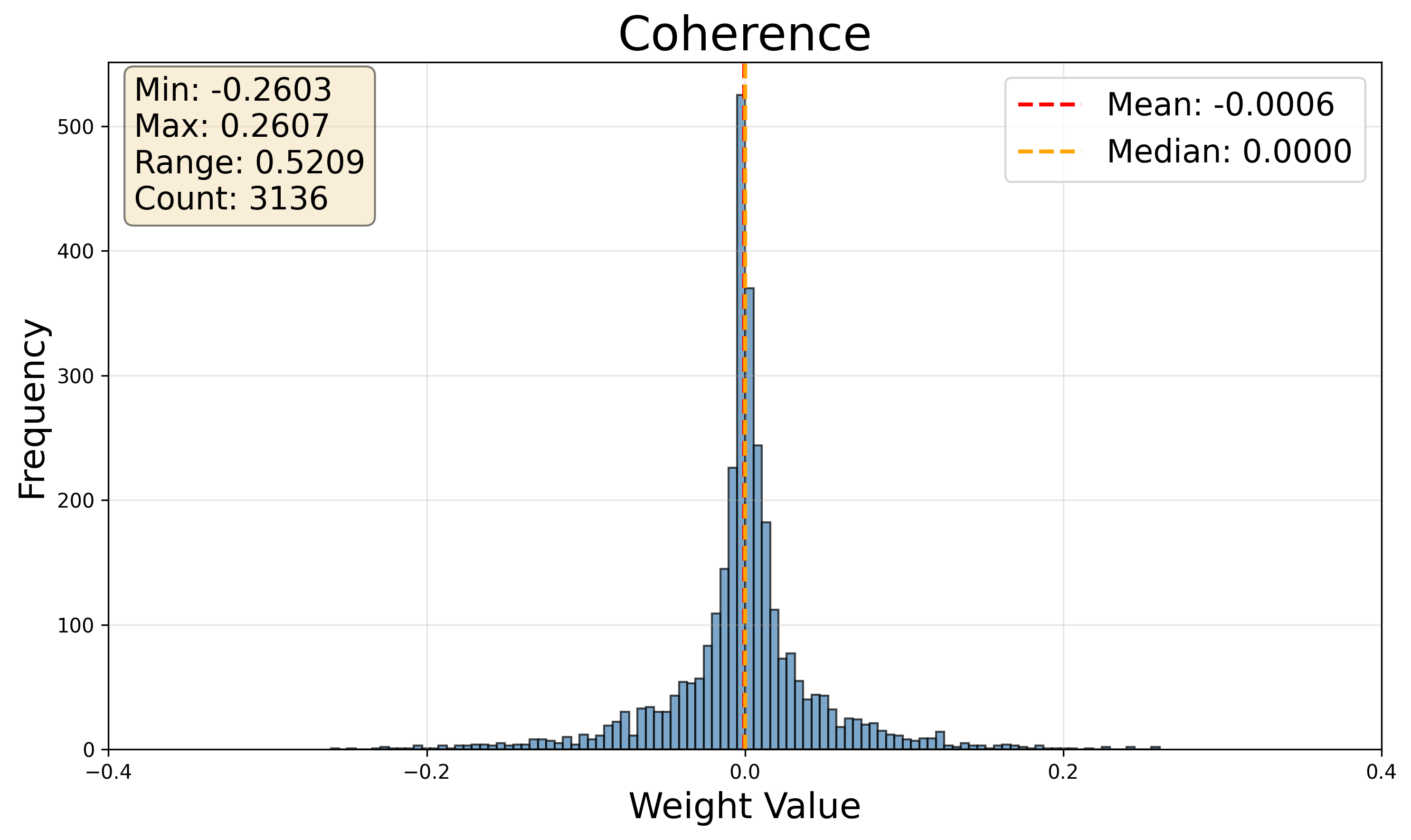}
    \caption{}
  \end{subfigure}\hfill
  \begin{subfigure}{0.3\textwidth}
    \includegraphics[width=\linewidth]{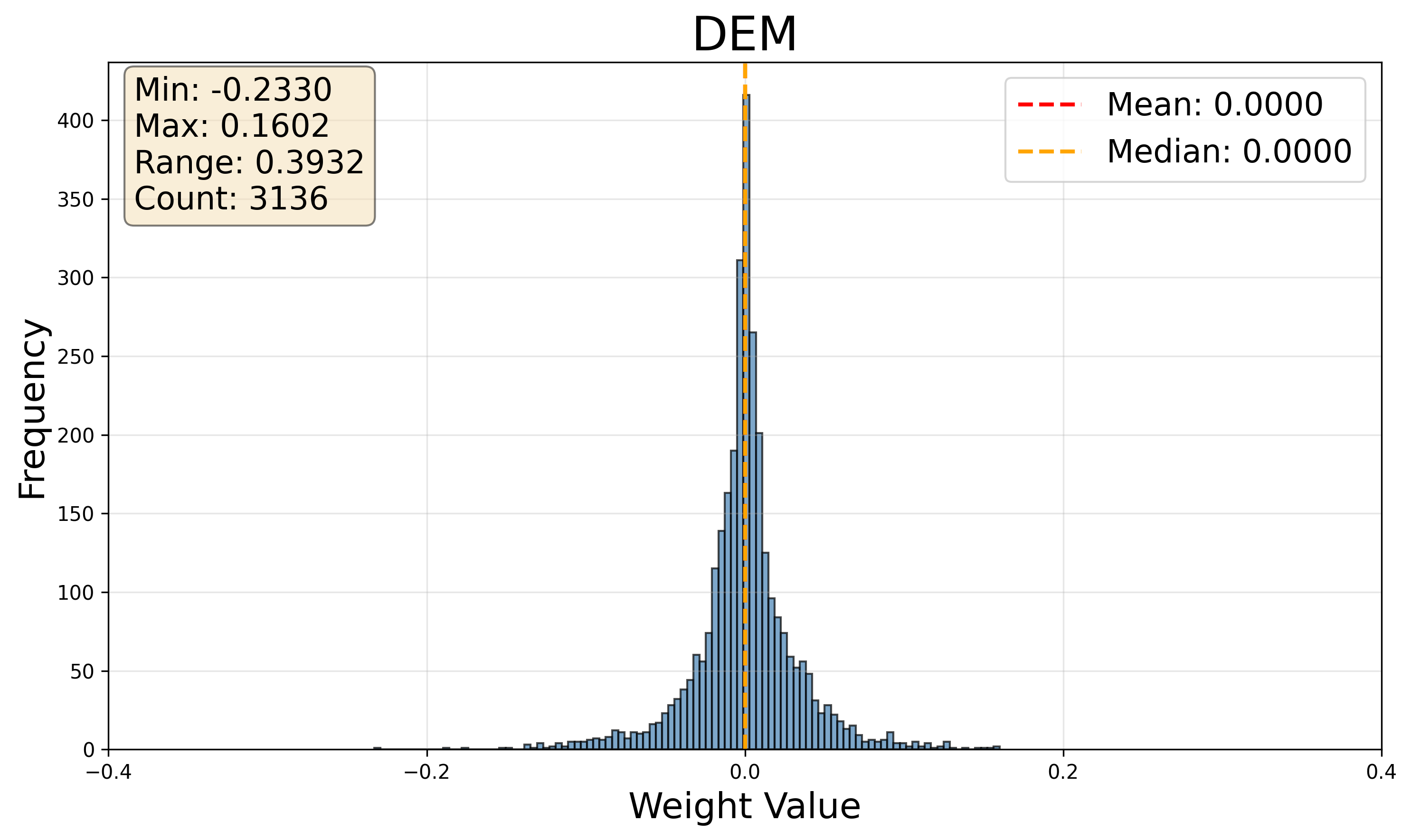}
    \caption{}
  \end{subfigure}
  
  \begin{subfigure}{0.3\textwidth}
    \includegraphics[width=\linewidth]{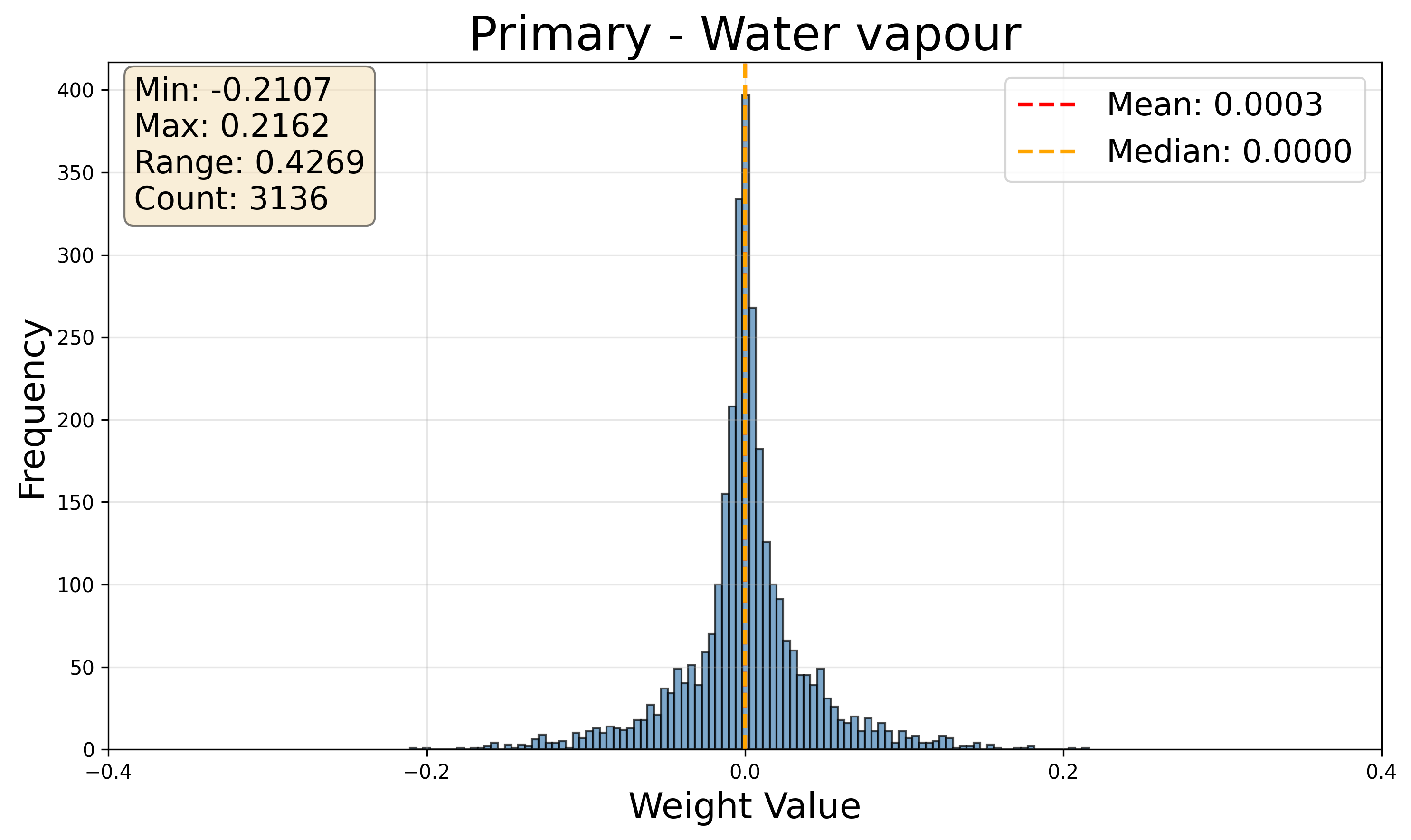}
    \caption{}
  \end{subfigure}\hfill
  \begin{subfigure}{0.3\textwidth}
    \includegraphics[width=\linewidth]{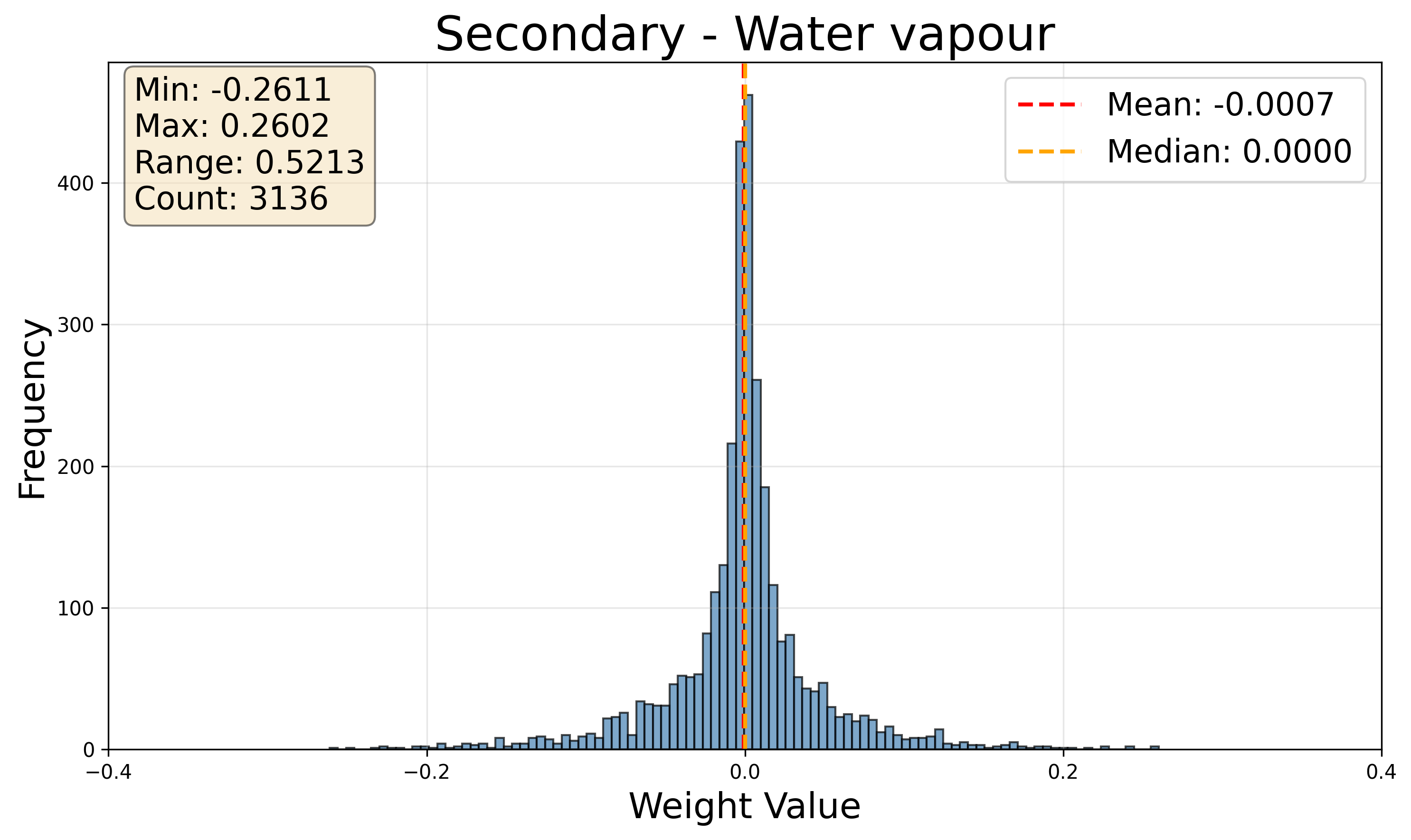}
    \caption{}
  \end{subfigure}\hfill
  \begin{subfigure}{0.3\textwidth}
    \includegraphics[width=\linewidth]{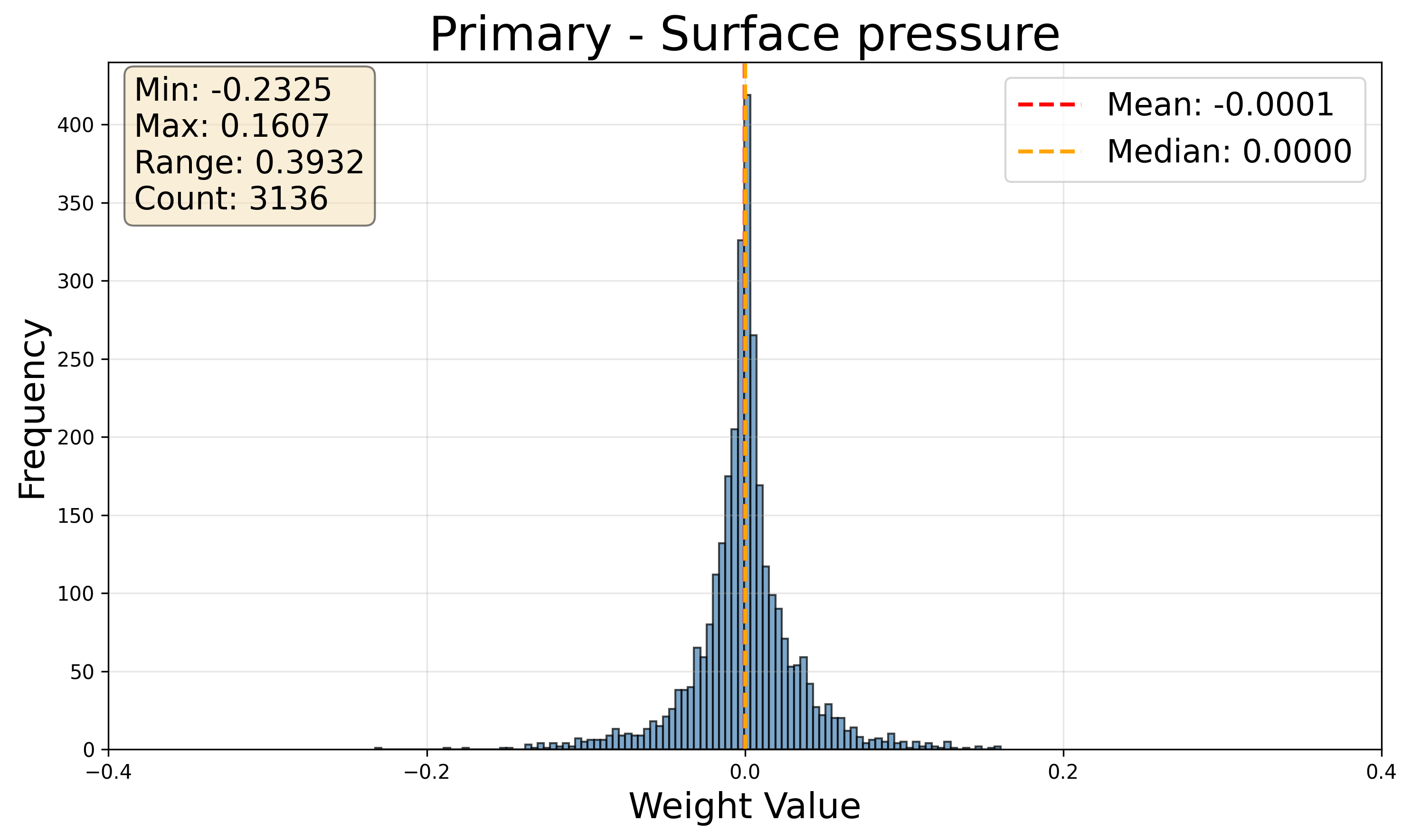}
    \caption{}
  \end{subfigure}
  
  \begin{subfigure}{0.3\textwidth}
    \includegraphics[width=\linewidth]{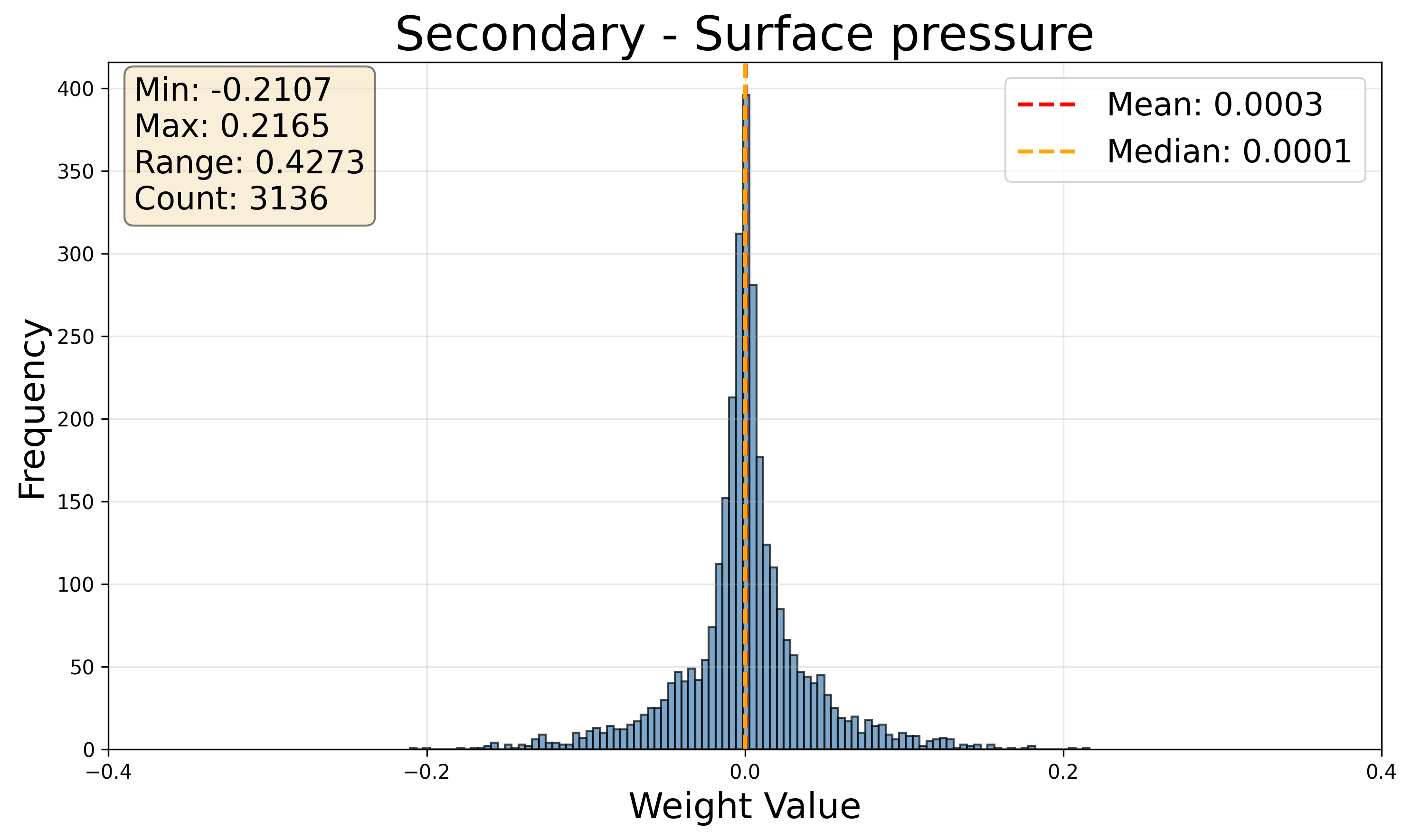}
    \caption{}
  \end{subfigure}\hfill
  \begin{subfigure}{0.3\textwidth}
    \includegraphics[width=\linewidth]{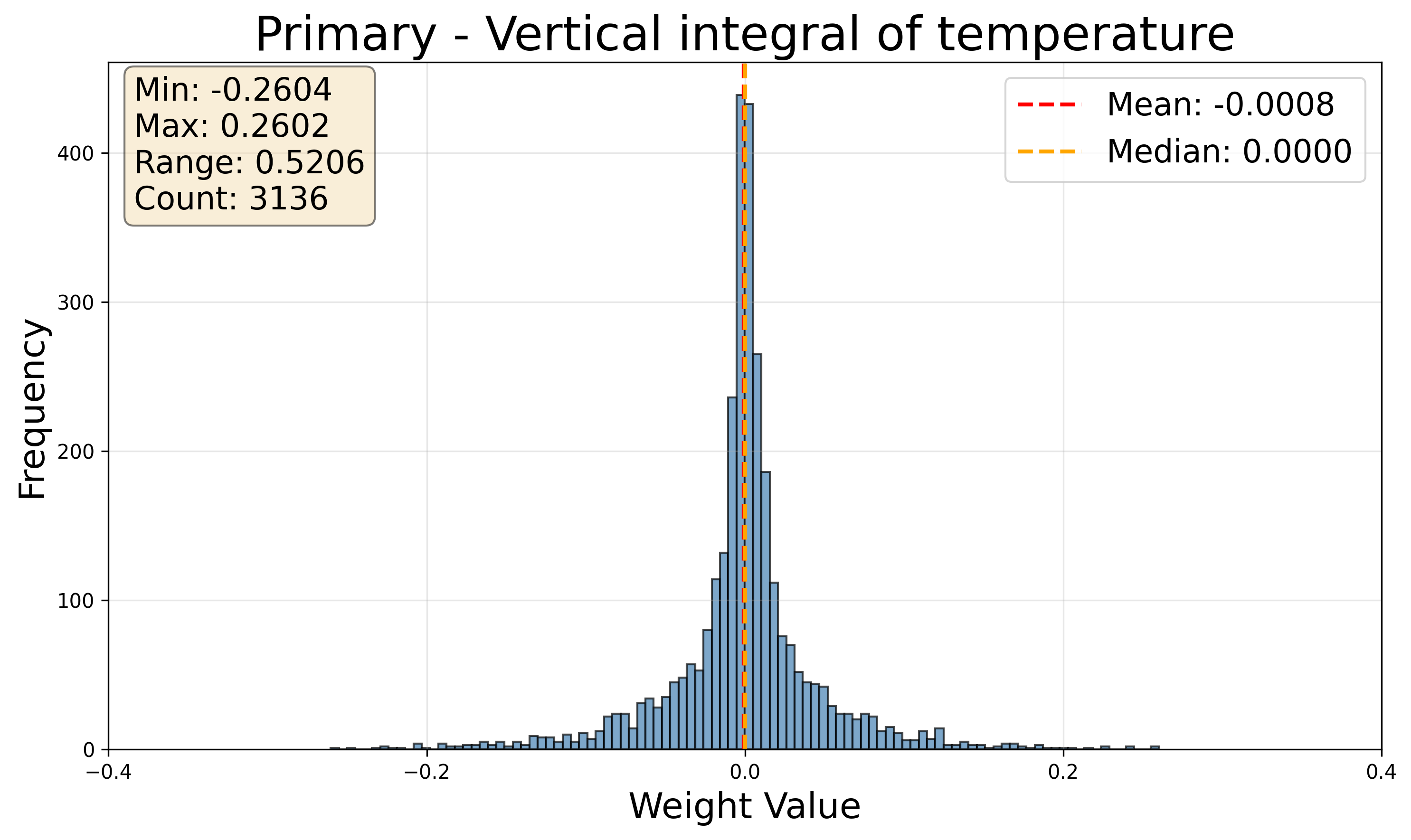}
    \caption{}
  \end{subfigure}\hfill
  \begin{subfigure}{0.3\textwidth}
    \includegraphics[width=\linewidth]{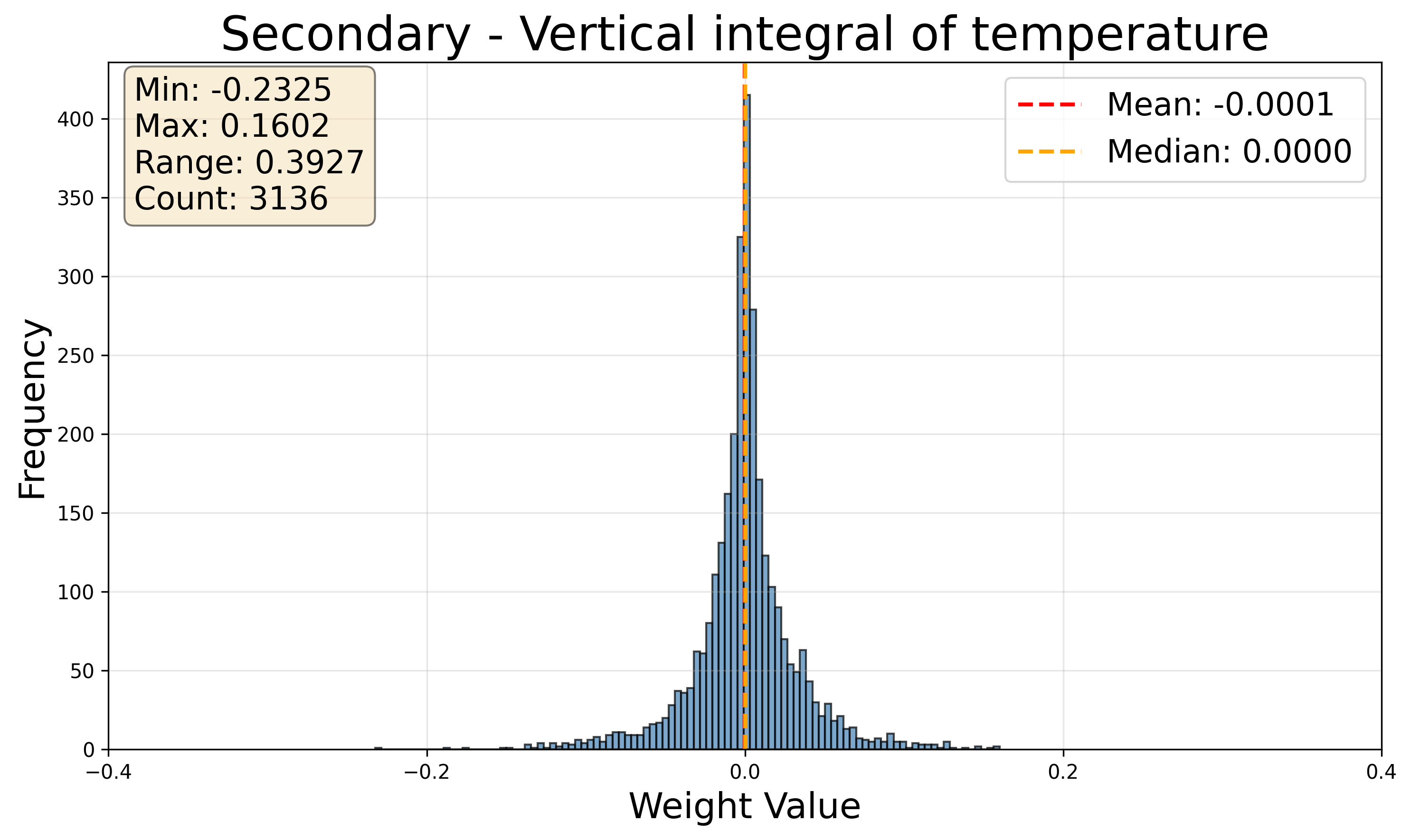}
    \caption{}
  \end{subfigure}
  
  \caption{Distributions of the convolutional weights for all nine input channels.}
  \label{fig:weights_9plots}
\end{figure*}

In early fusion all data sources are aggregated in the first convolutional layer resulting in $N$ output channels which will be subsequently used by the rest of the network. In contrast, in the base-setting all $N$ channels represent the high-resolution spatial modalities.  
To validate that atmospheric variables are utilized by the fusion layer, rather than being ignored, we examine the distribution of the kernel weights per input channel for the fusion layer of the best performing single-timestep model (see \cref{fig:weights_9plots}). We observe non-negligible weight magnitudes for atmospheric variables similar to imaging channels, indicating that atmospheric inputs actively contribute to the fused representation. 

\begin{table*}[ht]
\centering
\caption{Deformation segmentation metrics for the best-performing DeepLabV3 model (with atmospheric input) under different input masking conditions. Precision (Prec.), Recall (Rec.), F1-score (F1), and Intersection over Union (IoU) are reported for the deformation class.}
\label{tab:segm_mask_results}
\begin{tabular}{@{}llccccc@{}}
\toprule
\textbf{Setting} & \textbf{Model} & \textbf{Masked Input} & \textbf{Prec.} & \textbf{Rec.} & \textbf{F1} & \textbf{IoU} \\
\midrule
\multirow{3}{*}[-0.5em]{\textbf{Single-Timestep}} 
  & \multirow{3}{*}[-0.5em]{DeepLabv3} 
  & None
  & 81.64 
  & 60.82
  & 69.65 
  & 53.46 \\
\cmidrule{3-7}
 &
  & Atm.
  & 35.87 
  & 99.84
  & 50.48 
  & 33.76 \\
\cmidrule{3-7}
  &  
  & \begin{tabular}[c]{@{}l@{}}InSAR/DEM\end{tabular} 
  & ~~0.07 
  & ~~0.00 
  & ~~0.00 
  & ~~0.00 \\
\bottomrule
\end{tabular}
\end{table*}

To further examine the contribution of each modality, we conduct two ablation experiments, testing the model trained on the \textit{atmospheric-setting} with a) only atmospheric modalities and b) only InSAR, DEM and Coherence (masking the omitted modalities with zeros). As shown in \cref{tab:segm_mask_results}, in case a), we observe a total collapse ($0\%$ F1-Score), suggesting that atmospheric data alone, without InSAR information, cannot support segmentation, which is an expected result since the former do not provide any information on ground deformation processes. In case b), we observe degraded performance ($\approx 50\%$ F1-Score) compared to both the same model with full inputs ($\approx69.65\%$), and the base training setting ($\approx71.49\%$). This indicates that the model has learned to utilize the fusion of atmospheric information and the base-setting in a way that compromises its ability to extract all necessary information from the primary imaging modalities.

\bibliographystyle{ieeenat_fullname}
\bibliography{main}

\end{document}